\documentclass{article}

\usepackage[nonatbib, final]{timeseries_workshop}

\usepackage[utf8]{inputenc} %
\usepackage[T1]{fontenc}    %
\usepackage{hyperref}       %
\usepackage{url}            %
\usepackage{booktabs}       %
\usepackage{amsfonts}       %
\usepackage{nicefrac}       %
\usepackage{microtype}      %
\usepackage{xcolor}         %
\usepackage{graphicx}
\usepackage{arydshln}
\usepackage{multirow}
\usepackage{multicol}
\usepackage{subcaption}
\usepackage{wrapfig}
\usepackage[toc,page,header]{appendix}

\usepackage{minitoc}
\usepackage{arydshln}

\title{Pre-trained Forecasting Models: Strong Zero-Shot Feature Extractors for Time Series Classification}

\author{%
  Andreas Auer$^{\ 1,2}$ \ \ \
  Daniel Klotz$^{\ 3}$ \ \ \
  \textbf{Sebastian Böck}$^{\ 1}$ \ \ \
  \textbf{Sepp Hochreiter}$^{\ 1,2}$ \vspace{2mm} \\ 
{$^1$}{NXAI GmbH, Linz, Austria} \\
{$^2$}{ELLIS Unit, LIT AI Lab, Institute for Machine Learning, JKU Linz, Austria}\\
{$^3$}{Interdisciplinary Transformation University Austria, Linz, Austria}
}

\begin{document}

\maketitle
\doparttoc %
\faketableofcontents %

\begin{abstract}
Recent research on time series foundation models has primarily focused on forecasting, leaving it unclear how generalizable their learned representations are. 
In this study, we examine whether frozen pre-trained forecasting models can provide effective representations for classification.
To this end, we compare different representation extraction strategies and introduce two model-agnostic embedding augmentations. 
Our experiments show that the best forecasting models achieve classification accuracy that matches or even surpasses that of state-of-the-art models pre-trained specifically for classification. 
Moreover, we observe a positive correlation between forecasting and classification performance. 
These findings challenge the assumption that task-specific pre-training is necessary, and suggest that learning to forecast may provide a powerful route toward constructing general-purpose time series foundation models.
\end{abstract}

\section{Introduction}\label{sec:introduction}
In time series forecasting, foundation models are becoming increasingly prominent. 
They are large models that are pre-trained on broad data, and
therefore have the ability to generalize across unseen datasets \cite{ansariChronosLearningLanguage2024b, wooUnifiedTrainingUniversal2024a, dasDecoderonlyFoundationModel2024e, cohen2025timedifferentobservabilityperspective, auer2025tirex}. 
New benchmarks with public leaderboards such as 
GiftEval \cite{aksuGIFTEvalBenchmarkGeneral2024a} and 
BOOM \cite{cohen2025timedifferentobservabilityperspective} 
have accelerated advances in state-of-the-art methods.
Apart from forecasting, Time Series Classification (TSC) is another key application in time series analysis.

Earlier general-purpose time series models \cite{goswami2024moment, gao2024units} evaluated multiple downstream tasks,
but recent work shows that they have failed to reach state-of-the-art performance in either forecasting or classification \cite{feofanov2025mantis, ekambaramTinyTimeMixers2024b}. 
More recently, the majority of newly introduced 
foundation models \cite{auer2025tirex, cohen2025timedifferentobservabilityperspective, liu2025sundial, wang2025output, hoo2025tablestimetabpfnv2outperforms}
have been optimized specifically for forecasting,
and only few have focused on 
classification \cite{feofanov2025mantis, lin2024nutime}. 
Some argue that pre-training objectives should be aligned with downstream applications,
for example, contrastive objectives for classification or masked reconstruction 
for imputation \cite{feofanov2025mantis}. 
This perspective suggests that task-specialized pre-training may be necessary 
for optimal performance, which is in contrast to language and vision foundation models, 
where a single pre-trained model often transfers effectively across 
many diverse tasks \cite{bommasani2021opportunities, brownLanguageModelsAre2020b}.

This contrast motivates our central research question: 
\textbf{How well do representations from pre-trained forecasting models 
transfer to classification tasks?}
To answer this question, we evaluate a diverse set of forecasting models 
as frozen feature extractors on TSC benchmarks, 
analyze key design choices for representation extraction, and 
investigate the role of model architectures. 
Beyond the direct application to classification, 
our study aims to provide broader insights into 
the generalizability of learned representations, 
which is a step toward developing true time series foundation models.

Our contributions are as follows:
\textbf{(1)} We show that representations from pre-trained forecasting models yield classification accuracy on par with, and in some cases surpassing state-of-the-art models pre-trained explicitly for classification.
\textbf{(2)} We analyze design decisions for leveraging forecasting models in classification, providing practical guidance for future applications.
\textbf{(3)} We propose two model-agnostic representation augmentations that incorporate absolute statistical features and differentiated series to further improve classification performance.

The remainder of the paper introduces the problem setup, details our methodology for using forecasting models as feature extractors (Section~\ref{sec:method}), presents the experimental setup (Section~\ref{sec:experiments}) and results (Section~\ref{sec:results}), and concludes with key findings (Section~\ref{sec:conclusion}).

\paragraph{Problem Setup: Time Series Classification}

The TSC task is defined over a dataset $\mathcal{D} = \{(\mathbf{x}_i, y_i)\}_{i=1}^N$, where each sample consists of a time series $\mathbf{x}_i$ and its corresponding class label $y_i$.
A time series $\mathbf{x}_i \in \mathbb{R}^{T \times V}$ is a sequence of $T$ observations over $V$ variates, and the label $y_i$ belongs to one of $K$ discrete classes.
The objective is to learn a model that can accurately predict the label for a new, unseen time series. %

\section{Zero-Shot Forecasting Models as Classification Models}\label{sec:method}

We leverage pre-trained time series forecasting models as feature extractors.
Instead of training a classifier on the raw time series $\mathbf{x}_i$, we use a pre-trained model $E$ to map $\mathbf{x}_i$ to a latent representation $\mathbf{z}_i = E(\mathbf{x}_i)$, which is then fed into a simple classifier $C_L$ to output the final prediction $\hat{y}_i = C_L(\mathbf{z}_i)$. We refer to $\mathbf{z}_i$ also as \emph{embedding} of $\mathbf{x}_i$.

We exclusively use a zero-shot protocol for these models, meaning the parameters of the pre-trained model $E$ are frozen and never fine-tuned.
For each TSC dataset, we only train a standard out-of-the-box classifier $C_L$ on top of the embeddings produced by $E$.
This approach allows us to isolate and evaluate the quality and generalizability of the representations learned by the forecasting models.

\paragraph{Embedding Extraction \& Aggregation.}
Most state-of-the-art forecasting models do not specify a canonical method for extracting a single, fixed-size embedding for an entire time series.
However, as the majority utilize a transformer(-like)\footnote{TiRex~\cite{auer2025tirex} uses xLSTM \cite{beckXLSTMExtendedLong2024a} instead of a Transformer but still employs a block-based architecture \cite{beckXLSTM7BRecurrent2025}}, block-based architecture, we can extract hidden states at various points in the network.
This presents two key design choices:
how to aggregate information along (1) layer and (2) sequence dimensions.
We hypothesize that simply using the output from the final token of the final layer is suboptimal.
First, it is unclear which layer contains the best abstraction and transferable representation, as deeper layers often specialize to the original pre-training task, losing generalizability \cite{yosinski2014transferable, alkin2024mim}.
Second, relying on the last sequence position may neglect important information contained earlier in the series.

We investigate different aggregation strategies in our ablations.
For our main experiments, we apply mean pooling across the sequence dimension and concatenate these layer-wise representations.
This sequence-pooling strategy also inherently handles datasets with variable-length time series, ensuring a fixed-size embedding dimension.
The ablation study in Appendix~\ref{app:ext-results-agg-methods} confirms that aggregating across both dimensions is crucial.

\paragraph{Multivariate Data \& Univariate Models.}
Most top-performing pre-trained forecasting models are univariate.
For multivariate time‑series classification, we adopt a proven forecasting technique: treating each variate independently \cite{nieTimeSeriesWorth2022, auer2025tirex}.
We therefore process each of the $V$ variates independently through the frozen model $E$ to yield $V$ separate embeddings.

The subsequent design choice is how to aggregate these per-variate embeddings into a single representation.
We hypothesize that pooling discards variate-specific information, while concatenation preserves it.
Accordingly, we concatenate the per-variate embeddings in our main experiments, a choice empirically confirmed by our ablation studies (Appendix~\ref{app:ext-results-agg-methods}), which show concatenation consistently outperforms pooling.
We apply the same strategy to multivariate models that output per-variate embeddings.

\subsection{Embeddings Augmentations}

\paragraph{Absolute Sample Statistics.} A common characteristic of pre-trained forecasting models is the use of instance normalization.
While effective for forecasting, this removes all information regarding the absolute values and scale of the time series.
We hypothesize that for many classification tasks, this information might be an important discriminative signal.
To recover it, we propose to augment the model's embedding with basic sample statistics.
We divide the input time series $\mathbf{x}_i$ into $k$ non-overlapping patches ($k=8$ in our main experiments). 
For each patch, we calculate its mean, standard deviation, minimum, and maximum values. 
These statistics are then concatenated with the embedding $\mathbf{z}_i$ from the model to form the final representation.
Using a fixed number of patches ensures the resulting feature vector has a consistent size.

\paragraph{Time Series Differencing.}
Time series may contain strong trends that can dominate the signal and mask more subtle patterns.
To isolate these patterns, we propose to employ first-order differencing.
We generate a new, differenced time series by taking the difference between consecutive time steps ($\mathbf{x}'_t = \mathbf{x}_t - \mathbf{x}_{t-1}$).
This transformation, inspired by classical time series analysis, removes the local trend, making the resulting series more stationary and emphasizing step-to-step changes.
The differenced series is then processed by the same pre-trained model to produce a second embedding, which is concatenated to the original embedding.

\begin{figure}[t]
    \centering
    \small
    \begin{tabular}{lcccccccc}
\toprule
 & Type & ZS & \multicolumn{2}{c}{Univariate} & \multicolumn{2}{c}{Multivariate} & \multicolumn{2}{c}{Overall} \\
 &  &  & No Aug & Stat+Diff & No Aug & Stat+Diff & No Aug & Stat+Diff \\
\midrule
TiRex & Dec & yes & \textbf{0.80} & \textbf{0.81} & \textbf{0.74} & \textbf{0.74} & \textbf{0.79} & \textbf{0.80} \\
Chr. Bolt (Base) & EncDec & yes & 0.77 & 0.79 & 0.72 & \textbf{0.74} & 0.76 & 0.78 \\
Moirai (Large) & Enc & yes & 0.79 & 0.80 & 0.70 & 0.70 & 0.78 & 0.78 \\
TimesFM 2.0 & Dec & yes & 0.79 & 0.79 & 0.70 & 0.70 & 0.77 & 0.78 \\
TimesFM 1.0 & Dec & yes & 0.74 & 0.75 & 0.71 & 0.72 & 0.73 & 0.74 \\
Chronos (Base) & EncDec & yes & 0.71 & 0.76 & 0.71 & 0.72 & 0.71 & 0.75 \\
Toto & Dec & yes & 0.71 & 0.74 & 0.71 & 0.70 & 0.71 & 0.73 \\
\noalign{\vskip 1mm}\cdashline{1-1}\noalign{\vskip 1mm}
Mantis & Enc & no & \multicolumn{2}{c}{0.79} & \multicolumn{2}{c}{\textbf{0.74}}  & \multicolumn{2}{c}{0.78}  \\
NuTime & Enc & no & \multicolumn{2}{c}{0.67} &\multicolumn{2}{c}{0.68} & \multicolumn{2}{c}{0.67} \\
Moment (Large) & Enc & no & \multicolumn{2}{c}{0.63} & \multicolumn{2}{c}{0.57} & \multicolumn{2}{c}{0.62} \\
DTW & - & - & \multicolumn{2}{c}{0.73} & \multicolumn{2}{c}{0.72} & \multicolumn{2}{c}{0.73} \\
\bottomrule
\end{tabular}

    \captionof{table}{Classification accuracy of different models for the univariate, multivariate, and combined benchmark (Random Forest). ``Stat+Diff'' shows results with both proposed augmentations applied;  ``no Aug'' utilizes the pure forecasting model representations. ``ZS'' indicates models that did not have access to the benchmarks training data during pre-training.}
    \label{tab:main-result-table}
     \vspace{4mm}
    \centering
    \vspace{-2mm}
    \begin{subfigure}[t]{0.49\textwidth}
        \centering
        \includegraphics[width=\linewidth]{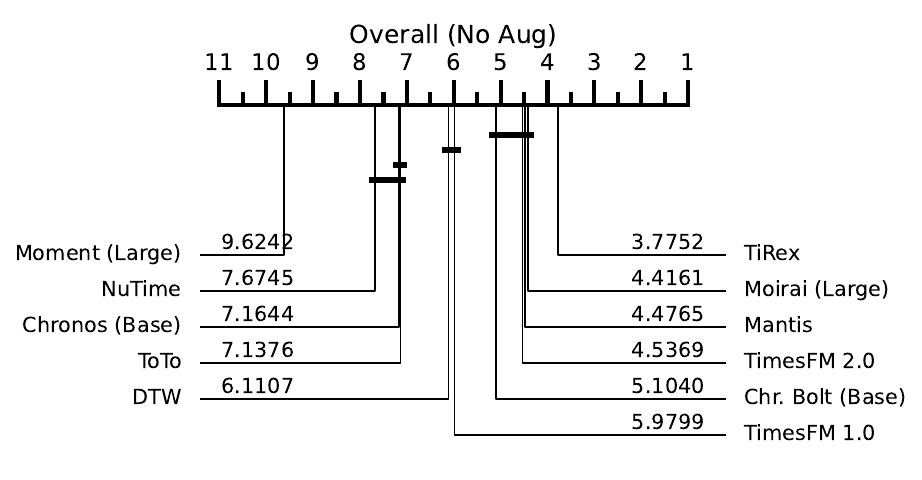}
        \label{fig:univariate_cd}
    \end{subfigure}
    \hfill
    \begin{subfigure}[t]{0.49\textwidth}
        \centering
        \includegraphics[width=\linewidth]{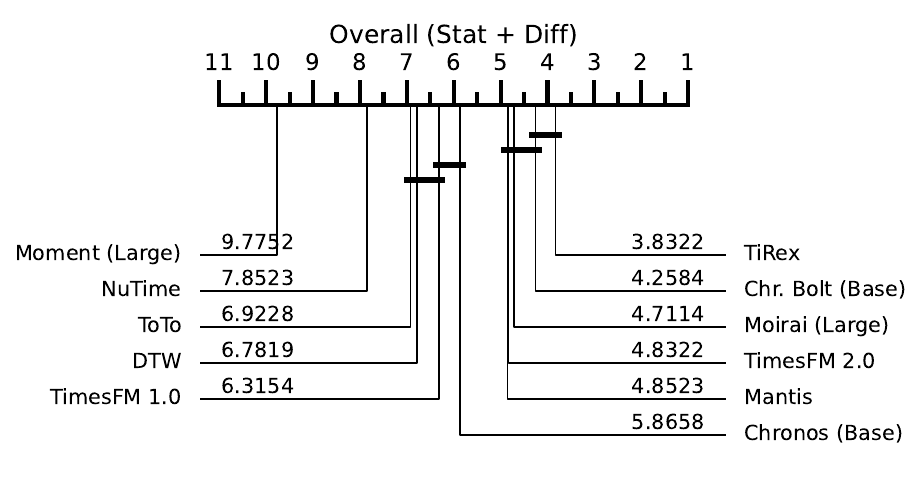}
        \label{fig:overall_cd}
    \end{subfigure}
    \vspace{-4mm}
    \captionof{figure}{Critical difference plot of the average accuracy ranks for the evaluated models across the combined benchmark datasets (Random Forest). Left without augmentation; right with augmentations.
    Models connected by a bar are not significantly different (Wilcoxon signed-rank test).}
    \label{fig:combined_analysis}
    \vspace{-2mm}
\end{figure}

\section{Experiments}\label{sec:experiments}

Our evaluation uses the UCR \cite{UCRArchive2018} and UEA \cite{bagnallUEAMultivariateTime2018} archives, comprising $127$ univariate and $30$ multivariate classification datasets with predefined train/test splits.
We excluded $5$ datasets with sample lengths exceeding $2048$ and $2$ others due to processing problems.
We evaluate a set of leading pre-trained forecasting models, including TiRex \cite{auer2025tirex}, Chronos (Bolt) \cite{ansariChronosLearningLanguage2024b}, TimesFM \cite{dasDecoderonlyFoundationModel2024e}, and Moirai \cite{wooUnifiedTrainingUniversal2024a} --- including the newest and previous model generations and different sizes.
These are compared against Moment \cite{goswami2024moment}, a ``general'' pre-trained model, the classification-specific pre-trained models NuTime \cite{lin2024nutime} and Mantis \cite{feofanov2025mantis}, and Dynamic Time Warping (DTW) \cite{berndt1994using} as a baseline.
For each pre-trained model, we extract embeddings and train a Random Forest, a linear layer, and a kNN classifier on top --- and evaluate accuracy.
Details on the experiment setup are presented in Appendix~\ref{app:experiment-details}.

\section{Results}\label{sec:results}

This section reports results using the best-performing classifier (Random Forest) and the largest model size for each model.
The main results are summarized in Table~\ref{tab:main-result-table} and Figure~\ref{fig:combined_analysis}.
Full results for all classifiers, model sizes, and ablations are available in Appendix~\ref{app:extended-results-toplevel}.
In the following, we discuss the individual aspects of our main findings.

\paragraph{Forecasting Models are Effective Zero-Shot Feature Extractors.}
The best forecasting models achieve accuracies competitive with or exceeding Mantis, a state-of-the-art model designed for this classification task.
This result is particularly interesting because the forecasting models had no exposure to the classification benchmarks during their pre-training, unlike Mantis and NuTime, which were also pre-trained on the training split of the benchmarks.
The results are robust across other classifier (Appendix~\ref{app:ext-results-classifier}), metrics (Appendix~\ref{app:ext-results-diffmetrcis}), and benchmark configuration (Appendix~\ref{app:ext-results-512length}).
This suggests that pre-training towards forecasting tasks might be a viable path for generating general-purpose time series representations.

\begin{wrapfigure}{r}{0.43\textwidth}
    \centering
    \vspace{-4mm}
    \includegraphics[width=0.43\textwidth]{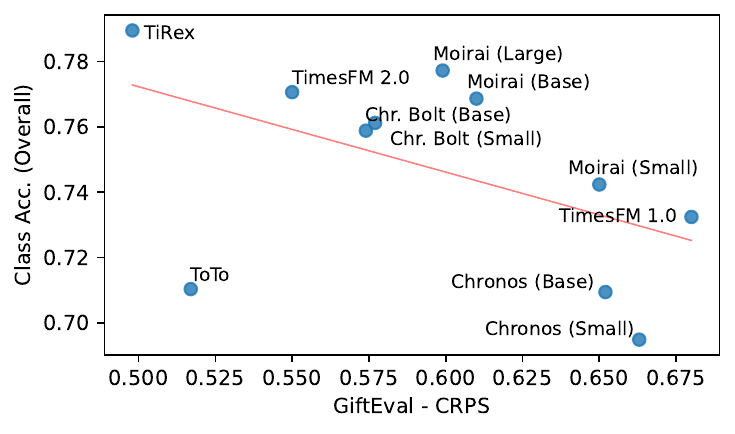}
    \caption{Classification accuracy versus forecasting performance (CRPS on GiftEval) of the evaluated models.
    The trend (red line) shows that better forecasting ability (lower CRPS) relates to higher classification accuracy.}
    \label{fig:gifteval-comparision}
\end{wrapfigure}

\paragraph{Forecasting and Classification Performance Correlate.}
We observe a positive correlation between a model's performance on the GiftEval forecasting benchmark \cite{aksuGIFTEvalBenchmarkGeneral2024a} and its classification accuracy (Figure~\ref{fig:gifteval-comparision}).
The trend has considerable noise, with notable under-performance from Chronos (potentially due to missing patch processing) and Toto.
However, overall this trend suggests that the features learned for accurate forecasting are transferable to classification tasks.

\paragraph{Impact of Model Architecture.}
Results do not point to a superior architectural paradigm (Encoder, Decoder, Encoder-Decoder).
Both, the top- and low-performing models, are diverse in that regard.
Regarding base architecture, TiRex, as the only non-Transformer model, performs best.
If the forecast advantage stems from its state-tracking capability, as prior work suggests \cite{auer2025tirex}, then this benefit seems to transfer to classification, implying a better general representation.

\paragraph{Efficacy of Augmentations.}
The proposed augmentations improve the results across most models --- the significance regarding the signed rank test varies between the results.
Detailed results including ablation of the individual augmentations and a qualitative analysis are presented in Appendix~\ref{app:ext-results-augment}.

\section{Conclusion}\label{sec:conclusion}
This work demonstrates that pre-trained forecasting models are effective zero-shot feature extractors for time series classification.
We found that representations from strong forecasting models match or even exceed the performance of specialized classification models --- particularly noteworthy as the forecasting models did not pre-train with benchmark training data, while the classification-specific models did.
This finding, combined with a positive correlation between forecasting and classification performance, questions the need for task-specific pre-training.

\paragraph{Limitations \& Future Work}
The work focuses on a zero-shot evaluation protocol and does not include fine-tuning.
This choice ensures a fair comparison of the base representations, as optimal fine-tuning strategies might be highly model-specific.
The work also omits a direct comparison to task-specific and supervised classifiers; however, prior work \cite{feofanov2025mantis, lin2024nutime, goswami2024moment} has already shown that the pre-trained classification models we evaluate are competitive with these.
Future work could probe the generalizability of these representations on other tasks, such as anomaly detection.

\begin{ack}
The ELLIS Unit Linz, the LIT AI Lab, and the Institute for Machine Learning are supported by the Federal State Upper Austria.
\end{ack}

\bibliographystyle{abbrv}

{
\small
\bibliography{references}

}

\clearpage

\appendix

{
\hypersetup{linkcolor=black}
\addcontentsline{toc}{section}{Appendix} %
\part{Appendix} %
\parttoc %
}

\section{Related Work}\label{app:related-work}

Pre-trained foundation models have become popular in time series analysis.
Early explorations adapted Large Language Model (LLM) for time series tasks \cite{gruver2023large}, while more recent models typically only borrow the architecture from LLM's but pre-train with time series tasks and data.
While there is a recent focus on forecasting \cite{rasul2024lagllama, wooUnifiedTrainingUniversal2024a, dasDecoderonlyFoundationModel2024e, ansariChronosLearningLanguage2024b, ekambaramTinyTimeMixers2024b, auer2025tirex, cohen2025timedifferentobservabilityperspective, liu2025sundial, hoo2025tablestimetabpfnv2outperforms, auer2025zero}, other literature has explored models for a wider range of downstream tasks, including classification:
General-purpose models like Moment \cite{goswami2024moment}, GPT4TS, \cite{zhou2023one}, and UniTS \cite{gao2024units} address classification alongside other tasks.
More specialized models, like Mantis \cite{feofanov2025mantis} and NuTime \cite{lin2024nutime} focus specifically on pre-training for classification tasks.
For our analysis, Moment, Mantis, and NuTime are particularly suitable as they allow feature extraction without task-specific fine-tuning.
We note, however, that they are not ``zero-shot'' on our evaluated benchmark, as their pre-training corpora include the training split of the benchmark.

Distinct from the generalizable pre-training paradigm, another line of research involves task-specific unsupervised classification models.
These methods are typically trained per-dataset.
While some support limited transfer learning, they do not allow for zero-shot feature extraction with a single, fixed model.
Notable examples include TLoss \cite{Franceschi2019TLoss}, TS2Vec\cite{yue2022ts2vec}, TF-C \cite{zhang2022selfsupervised} and Ti-MAE \cite{li2023ti}.

Additionally, there is extensive literature on supervised classification models that are mostly not based on deep learning. 
These classical methods often rely on ensembles and heuristically engineered features.
\cite{middlehurstBakeReduxReview2024} and \cite{ruiz2021great} provide a good overview of these methods.

\clearpage

\section{Experiment Details}\label{app:experiment-details}

\subsection{Benchmark Data}

For our evaluation we utilize the UCR \cite{UCRArchive2018} (127 univariate datasets) and the UEA \cite{bagnallUEAMultivariateTime2018} (30 multivariate datasets) classification benchmark datasets.
The benchmark covers various types and domains of time series including for example sensor, audio, motion or health data.
The train and test split is predefined by the benchmarks.
We removed datasets with a sample length over $2048$, which is the case for $5$ datasets.
Specifically, these are: \emph{MotorImagery, HandOutlines, StandWalkJump, EigenWorms}, and \emph{Rock}.
Further, we removed \emph{InsectWingbeat} and \emph{PLAID} as these lead to processing problems across the majority of models in the classifier training, likely due to their size.

\subsection{Pre-trained Models and Implementation Details}

We evaluate a suite of prominent pre-trained forecasting models:
TiRex \cite{auer2025tirex},
ToTo \cite{cohen2025timedifferentobservabilityperspective}, Chronos \cite{ansariChronosLearningLanguage2024b},
Chronos Bolt \cite{ansariabdulfatirFastAccurateZeroshot2024}, 
TimesFM (1.0 and 2.0) \cite{dasDecoderonlyFoundationModel2024e},
and Moirai\cite{wooUnifiedTrainingUniversal2024a}.
When possible (e.g.,\ for TimesFM and Chronos), we analyze both the newest and the previous generation of the models.
This gives a better insight into how improvements in forecasting translate to gains in classification, i.e., how they reflect enhancements in the general, underlying representation.
We compare these forecasting models to Moment~\cite{goswami2024moment}, NuTime~\cite{lin2024nutime}, and Mantis~\cite{feofanov2025mantis}.
Moment is a "general" pre-trained time series models -- NuTime and Mantis are classification-specific pre-trained models.
These models support a feature extraction approach as introduced in Section~\ref{sec:method} without fine-tuning.
However, they are not really ``zero-shot'' as (parts) of the training data of the classification benchmark are utilized in pre-training.
Additionally, we compare to Dynamic Time Warping (DTW) as a baseline.
Implementation details are provided in the following:

\begin{itemize}
    \item \textbf{TiRex} \cite{auer2025tirex}: We utilize the official pre-trained weights from \href{https://huggingface.co/NX-AI/TiRex}{Hugginface} and adapt the original source code from \href{https://github.com/NX-AI/tirex}{GitHub} to extract hidden layer representations.
    \item \textbf{Chronos / Chronos Bolt} \cite{ansariChronosLearningLanguage2024b, ansariabdulfatirFastAccurateZeroshot2024}: For both Chronos and Chronos Bolt, we evaluate the small and base model size.
    This model family is unique in providing a dedicated API for embedding extraction.
    We utilize this API, which returns a single-layer representation, and therefore only perform aggregation along the sequence dimension.
    \item \textbf{TimesFM} \cite{dasDecoderonlyFoundationModel2024e}:For both versions 1.0 and 2.0, we use the official PyTorch weights from Hugging Face and modify the source code from \href{https://github.com/google-research/timesfm/tree/master/notebooks}{GitHub} to access hidden states from all decoder layers.
    \item \textbf{Moirai} \cite{wooUnifiedTrainingUniversal2024a}: We evaluate Moirai 1.1 in all model sizes (small, base, and large).
    We utilize the official pre-trained weights from Hugginface and adapt the original source code from \href{https://github.com/SalesforceAIResearch/uni2ts}{GitHub} to extract hidden layer representations.
    While inherently multivariate, Moirai's ``variate flattening'' fails on datasets with a very high number of variates due to memory constraints.
    In these cases, we apply the model in a univariate fashion to each variate and concatenate the resulting embeddings.
    This is the case for the following datasets: 
    \begin{itemize}
        \item Moirai Small: \emph{FaceDetection, Heartbeat, MotorImagery, PEMS-SF, SpokenArabicDigits}
        \item Moirai Base: \emph{FaceDetection, Heartbeat, MotorImagery, PEMS-SF, PhonemeSpectra, SpokenArabicDigits}
        \item Moirai Large: \emph{FaceDetection, FingerMovements, Heartbeat, LSST, MotorImagery, NATOPS, PEMS-SF, PhonemeSpectra}
    \end{itemize}
    \item \textbf{Mantis} \cite{feofanov2025mantis}: We follow the official zero-shot feature extraction procedure from their \href{https://github.com/vfeofanov/mantis/blob/main/getting_started/single_channel_extract_feats.ipynb}{Github repository}, which includes interpolating all time series to a fixed length of 512 before embedding.
    \item \textbf{NuTime} \cite{lin2024nutime}: Following the protocol in \cite{feofanov2025mantis} we use the pre-trained weights provided in the respective \href{https://github.com/chenguolin/NuTime}{GitHub repository}, while utilizing the hyperparameters according to \href{https://github.com/chenguolin/NuTime/blob/main/configs/demo_ft_epilepsy.json}{this configuration file}.
    We use NuTime in zero-shot feature extraction mode, i.e., variates are embedded independently.
    \item \textbf{Moment} \cite{goswami2024moment}: We use the official zero-shot feature extraction method as demonstrated in their \href{https://github.com/moment-timeseries-foundation-model/moment/blob/main/tutorials/representation_learning.ipynb}{GitHub repository} and evaluate all size variants (small, base, and large).
    \item \textbf{Dynamic Time Warping (DTW)}: We use the implementation of the aeon library \cite{middlehurst2024aeon}.
\end{itemize}

\subsection{Failure Fallback: DTW}

Certain model and dataset combinations result in computational failures (e.g., out-of-memory errors). 
To avoid skewing aggregate metrics by either dropping these results or assigning a score of zero, we adopt a fallback strategy:
For any failed run, we substitute the model's result with the performance of our DTW baseline on that specific dataset.
This approach ensures a complete comparison, mirroring a practical scenario.
Fallbacks were utilized for the following model-dataset combinations:

\begin{itemize}
    \item TimesFM 1.0: \emph{Crop, FaceDetection}
    \item TimesFM 2.0: \emph{FaceDetection, PEMS-SF, SpokenArabicDigits}
    \item Moirai (Large): \emph{Crop, ElectricDevices, StarLightCurves, PenDigits, SpokenArabicDigits}
    \item Moment (Base \& Large): \emph{PEMS-SF}
\end{itemize}

\subsection{Classifier Training \& Hyperparameter}

We evaluate three classifiers on the extracted embeddings:
Random Forest (as suggested by \cite{feofanov2025mantis} for Mantis), a linear model, and kNN as a baseline.
This tests the linear and non-linear separability of the embeddings.
Details are provided in the following:

\begin{itemize}
    \item \textbf{Random Forest} implemented with scikit-learn \cite{pedregosa11a2011sklearn}.
    Following the protocol from \cite{feofanov2025mantis}, we use ``n\_estimators=300''; keeping all other parameters at their default values.
    \item \textbf{Linear Model} implemented with PyTorch. It consists of a single linear layer trained with the AdamW optimizer (learning rate $10^{-4}$, weight decay $10^{-2}$). We use a 20\% validation split from the training data for early stopping (patience of 100), with a maximum of 10,000 epochs.
    \item \textbf{kNN} implemented with scikit-learn \cite{pedregosa11a2011sklearn}. We use ``n\_neighbors=1'' (1-NN) with the cosine similarity as distance metric.
\end{itemize}

\subsection{Critical Difference Plots}

All critical difference plots in the paper show the average accuracy rank of each method (lower is better).
A horizontal bar connects models with no statistically significant difference in performance.
This significance is determined by a pairwise Wilcoxon signed-rank test with a Holm correction at a significance level of $\alpha=0.1$.

\clearpage

\section{Extended Results}\label{app:extended-results-toplevel}

This section provides extended results to Section~\ref{sec:experiments}.
Extending Table~\ref{tab:main-result-table},
Table~\ref{tab:main-table-ext} shows the results for all evaluated model sizes.
In almost all cases, larger models perform better, which aligns with the performance trend observed in forecasting.
A notable exception is Moment, where the base model outperforms the large version.

The following subsections provide further analysis, including results for different classifiers (Section~\ref{app:ext-results-classifier}), ablations of the aggregation methods (Section~\ref{app:ext-results-agg-methods}), ablations and analysis of the embedding augmentation (Section~\ref{app:ext-results-augment}), and robustness checks using a different metric (Section~\ref{app:ext-results-diffmetrcis}) and a dataset subset (Section~\ref{app:ext-results-512length}).
The main results for each individual dataset are presented in Table~\ref{tab:individual-part1} - \ref{tab:individual-part6}.

\begin{table}
    \centering
    \begin{tabular}{lcccccccc}
\toprule
 & Type & ZS & \multicolumn{2}{c}{Univariate} & \multicolumn{2}{c}{Multivariate} & \multicolumn{2}{c}{Overall} \\
 &  &  & No Aug & Stat+Diff & No Aug & Stat+Diff & No Aug & Stat+Diff \\
\midrule
TiRex & Dec & yes & \textbf{0.80} & \textbf{0.81} & \textbf{0.74} & \textbf{0.74} & \textbf{0.79} & \textbf{0.80} \\
Chr. Bolt (Base) & EncDec & yes & 0.77 & 0.79 & 0.72 & \textbf{0.74} & 0.76 & 0.78 \\
Chr. Bolt (Small) & EncDec & yes & 0.77 & 0.79 & 0.73 & \textbf{0.74} & 0.76 & 0.78 \\
Moirai (Large) & Enc & yes & 0.79 & 0.80 & 0.70 & 0.70 & 0.78 & 0.78 \\
Moirai (Base) & Enc & yes & 0.79 & 0.79 & 0.69 & 0.71 & 0.77 & 0.78 \\
Moirai (Small) & Enc & yes & 0.75 & 0.77 & 0.69 & 0.73 & 0.74 & 0.77 \\
TimesFM 2.0 & Dec & yes & 0.79 & 0.79 & 0.70 & 0.70 & 0.77 & 0.78 \\
TimesFM 1.0 & Dec & yes & 0.74 & 0.75 & 0.71 & 0.72 & 0.73 & 0.74 \\
Chronos (Base) & EncDec & yes & 0.71 & 0.76 & 0.71 & 0.72 & 0.71 & 0.75 \\
Chronos (Small) & EncDec & yes & 0.70 & 0.75 & 0.70 & 0.72 & 0.70 & 0.75 \\
ToTo & Dec & yes & 0.71 & 0.74 & 0.71 & 0.70 & 0.71 & 0.73 \\
\noalign{\vskip 1mm}\cdashline{1-1}\noalign{\vskip 1mm}
Mantis & Enc & no & \multicolumn{2}{c}{0.79} & \multicolumn{2}{c}{\textbf{0.74}} & \multicolumn{2}{c}{0.78} \\
NuTime & Enc & no & \multicolumn{2}{c}{0.67} & \multicolumn{2}{c}{0.68} & \multicolumn{2}{c}{0.67} \\
Moment (Large) & Enc & no & \multicolumn{2}{c}{0.63} & \multicolumn{2}{c}{0.57} & \multicolumn{2}{c}{0.62} \\
Moment (Base) & Enc & no & \multicolumn{2}{c}{0.65} & \multicolumn{2}{c}{0.57} & \multicolumn{2}{c}{0.64} \\
Moment (Small) & Enc & no & \multicolumn{2}{c}{0.63} & \multicolumn{2}{c}{0.56} & \multicolumn{2}{c}{0.62} \\
DTW (1-NN) & - & - & \multicolumn{2}{c}{0.73} & \multicolumn{2}{c}{0.72} & \multicolumn{2}{c}{0.73} \\
DTW (3-NN) & - & - & \multicolumn{2}{c}{0.71} & \multicolumn{2}{c}{0.71} & \multicolumn{2}{c}{0.71} \\
\bottomrule
\end{tabular}

    \caption{Classification Accuracy of different models (and sizes) for the univariate, multivariate, and combined benchmark (Random Forest). ``Stat+Diff'' shows results with both proposed augmentations applied;  ``no Aug'' utilizes the pure forecasting model representations. ``ZS'' indicates models that did not have access to the benchmarks training data during pre-training.}
    \label{tab:main-table-ext}
\end{table}

\begin{table}[]
    \centering
    \begin{tabular}{lcccccc}
\toprule
 & \multicolumn{3}{c}{Linear} & \multicolumn{3}{c}{1-NN} \\
 & Univariate & Multivariate & Overall & Univariate & Multivariate & Overall \\
\midrule
TiRex & 0.78 & 0.72 & \textbf{0.77} & 0.75 & 0.67 & 0.74 \\
Chr. Bolt (Base) & 0.76 & \textbf{0.73} & 0.76 & 0.75 & 0.68 & 0.74 \\
Chr. Bolt (Small) & 0.76 & \textbf{0.73} & 0.75 & 0.75 & 0.68 & 0.74 \\
Moirai (Large) & \textbf{0.79} & 0.70 & \textbf{0.77} & \textbf{0.77} & 0.64 & 0.75 \\
Moirai (Base) & 0.78 & 0.70 & 0.76 & 0.76 & 0.65 & 0.74 \\
Moirai (Small) & 0.75 & 0.69 & 0.74 & 0.72 & 0.63 & 0.71 \\
TimesFM 2.0 & 0.75 & 0.70 & 0.74 & 0.71 & 0.56 & 0.69 \\
TimesFM 1.0 & 0.73 & 0.69 & 0.72 & 0.70 & 0.65 & 0.69 \\
Chronos (Base) & 0.71 & 0.72 & 0.71 & 0.67 & 0.66 & 0.67 \\
Chronos (Small) & 0.69 & 0.70 & 0.69 & 0.66 & 0.66 & 0.66 \\
ToTo & 0.70 & 0.71 & 0.70 & 0.65 & 0.63 & 0.65 \\
\noalign{\vskip 1mm}\cdashline{1-1}\noalign{\vskip 1mm}
Mantis & 0.77 & \textbf{0.73} & 0.76 & \textbf{0.77} & \textbf{0.72} & \textbf{0.76} \\
NuTime & 0.59 & 0.63 & 0.59 & 0.60 & 0.61 & 0.60 \\
Moment (Large) & 0.58 & 0.44 & 0.55 & 0.61 & 0.55 & 0.60 \\
Moment (Base) & 0.58 & 0.48 & 0.56 & 0.56 & 0.50 & 0.55 \\
Moment (Small) & 0.54 & 0.47 & 0.53 & 0.53 & 0.48 & 0.52 \\
DTW (1-NN) & 0.73 & 0.72 & 0.73 & 0.73 & \textbf{0.72} & 0.73 \\
DTW (3-NN) & 0.71 & 0.71 & 0.71 & 0.71 & 0.71 & 0.71 \\
\bottomrule
\end{tabular}

    \caption{Classification accuracy of different models and classifiers (linear model and 1-NN) for the univariate, multivariate, and combined benchmark.}
    \label{tab:other-classifier}
\end{table}

\subsection{Results for different Classifiers}\label{app:ext-results-classifier}

This section complements the main paper's evaluation by presenting the results for the other two classifiers:
the gradient-based trained linear model and the 1-NN baseline.
The results are shown in Table~\ref{tab:other-classifier}.

The overall performance ranking of the models is largely consistent with the main evaluation, which uses a Random Forest.
While there are minor shifts in relative performance --- for example, with the linear classifier, the results for TiRex and Chronos-Bolt are not significantly different --- key insights from our paper hold.
The best forecasting models perform on par with pre-trained classification models and forecasting performance is correlated with classification accuracy.
However, a difference is that when using the simplest classifier (1-NN), the forecasting models no longer outperform Mantis, the best pre-trained classification model.

We hypothesize that this discrepancy arises because less powerful classifiers, such as linear models or kNN, have a limited ability to transform the feature space.
The embedding space of a model pre-trained on classification, like Mantis, might be already better aligned with the classification task.
In contrast, a non-linear model like a Random Forest can better identify and exploit the relevant discriminative information, which we assume is present in the embeddings from both forecasting and classification models.

\clearpage

\subsection{Ablation Analysis: Aggregation Methods}\label{app:ext-results-agg-methods}

\paragraph{Sequence \& Layer Aggregation}
We conduct an ablation study of the method to aggregate the hidden states across both the layer and sequence dimensions.
For layer aggregation, we evaluated four strategies: concatenation of all layer representations, mean pooling, max pooling, and using only the representation from the last layer.
For sequence aggregation, we considered mean pooling, max pooling, and using the last output.
Concatenation is not a viable option for the sequence dimension, as it would result in variable-length embeddings dependent on the sample length.
After the sequence aggregation and before the layer aggregation we normalize the embeddings as different layers might operate in different feature spaces.
For the Chronos models, which have a predefined method for embedding extraction, we only ablated the sequence aggregation strategy.

Figures~\ref{fig:ablation-agg-tirex}-\ref{fig:ablation-agg-chronossmall} present the results.
Each figure presents a table with the mean accuracy over univariate, multivariate, and all datasets, complemented by a critical difference plot of mean ranks to visualize statistical significance.
Across almost all models, the combination of mean pooling over the sequence dimension and concatenation over the layer dimension is the top-performing strategy.
In no case any other strategy combination performs significantly better.

\newcommand{\aggcaption}[1]{%
    Results for \textbf{#1} for the \textbf{layer and sequence aggregation ablation} experiments. 
    (a) Average accuracy on univariate (Uni), multivariate (Multi), and overall (Comb) benchmark datasets. Sorted by overall accuracy. 
    (b) Critical difference diagram of the average accuracy ranks.%
}

    \begin{figure}[htbp]
        \centering
        \subcaptionbox{Average Accuracy}{%
            \begin{minipage}{0.4\textwidth}
                \centering
                \small
                \begin{tabular}{llrrr}
\toprule
 &  & Uni & Multi & Comb \\
Seq & Layer &  &  &  \\
\midrule
Mean & Concat & \textbf{0.80} & \textbf{0.74} & \textbf{0.79} \\
Mean & Mean & 0.79 & 0.73 & 0.78 \\
Max & Concat & 0.79 & 0.73 & 0.78 \\
Mean & Max & 0.79 & 0.72 & 0.77 \\
Max & Mean & 0.77 & 0.73 & 0.77 \\
Max & Max & 0.76 & 0.72 & 0.75 \\
Last & Concat & 0.75 & 0.73 & 0.75 \\
Mean & Last & 0.75 & 0.71 & 0.75 \\
Last & Mean & 0.74 & 0.72 & 0.74 \\
Last & Max & 0.73 & 0.71 & 0.73 \\
Max & Last & 0.72 & 0.70 & 0.72 \\
Last & Last & 0.70 & 0.69 & 0.70 \\
\bottomrule
\end{tabular}

            \end{minipage}
        }%
        \hfill %
        \subcaptionbox{Average Accuracy Rank (Overall benchmark)}{%
            \includegraphics[width=0.58\textwidth]{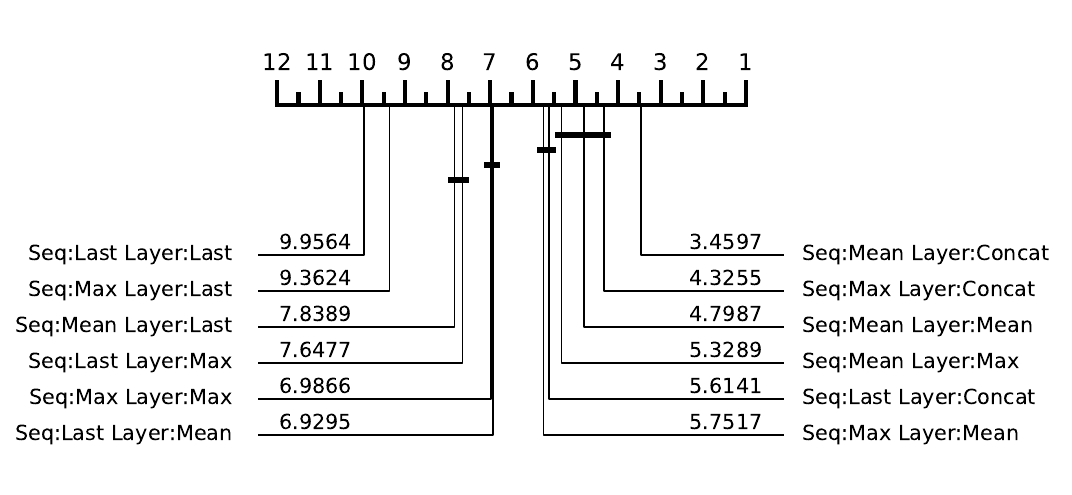}
        }
        \caption{\aggcaption{TiRex}}
        \label{fig:ablation-agg-tirex}
    \end{figure}

    \begin{figure}[htbp]
        \centering
        \subcaptionbox{Average Accuracy}{%
            \begin{minipage}{0.4\textwidth}
                \centering
                \small
                \begin{tabular}{llrrr}
\toprule
 &  & Uni & Multi & Comb \\
seq & L &  &  &  \\
\midrule
Mean & Concat & \textbf{0.79} & 0.69 & \textbf{0.77} \\
Max & Concat & 0.78 & \textbf{0.70} & 0.76 \\
Mean & Mean & 0.78 & 0.69 & 0.76 \\
Max & Mean & 0.77 & 0.69 & 0.76 \\
Mean & Max & 0.77 & 0.68 & 0.76 \\
Last & Concat & 0.77 & 0.69 & 0.76 \\
Mean & Last & 0.77 & 0.68 & 0.75 \\
Last & Mean & 0.76 & 0.68 & 0.75 \\
Max & Max & 0.76 & 0.68 & 0.75 \\
Last & Max & 0.75 & 0.67 & 0.74 \\
Max & Last & 0.75 & 0.67 & 0.73 \\
Last & Last & 0.73 & 0.64 & 0.71 \\
\bottomrule
\end{tabular}

            \end{minipage}
        }%
        \hfill %
        \subcaptionbox{Average Accuracy Rank (Overall benchmark)}{%
            \includegraphics[width=0.58\textwidth]{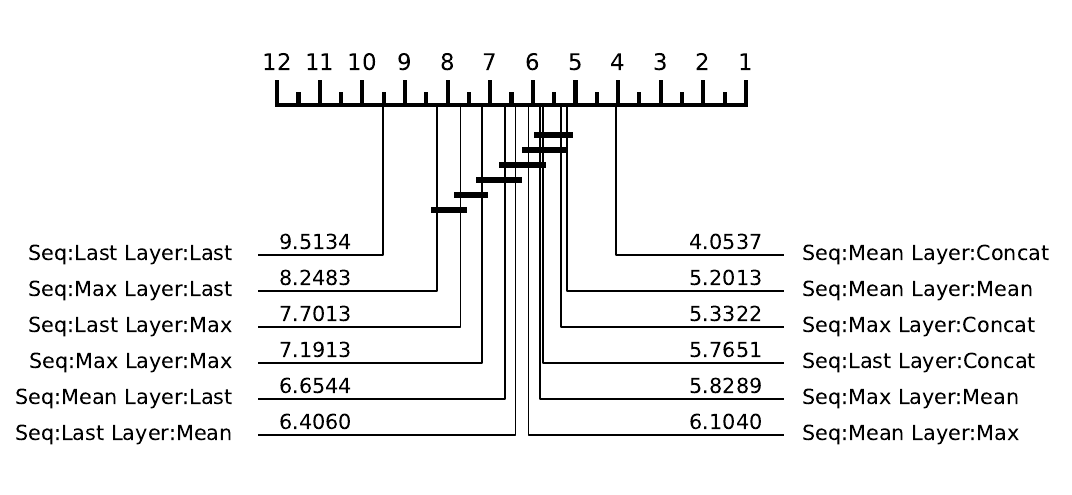}
        }
        \caption{\aggcaption{Moirai 1.1 (Base)}}
        \label{fig:ablation-agg-moiraibase11}
    \end{figure}

    \begin{figure}[htbp]
        \centering
        \subcaptionbox{Average Accuracy}{%
            \begin{minipage}{0.4\textwidth}
                \centering
                \small
                \begin{tabular}{llrrr}
\toprule
 &  & Uni & Multi & Comb \\
Seq & Layer &  &  &  \\
\midrule
Mean & Concat & \textbf{0.79} & 0.70 & \textbf{0.77} \\
Max & Concat & 0.78 & 0.71 & \textbf{0.77} \\
Last & Concat & 0.77 & \textbf{0.74} & \textbf{0.77} \\
Max & Mean & 0.77 & 0.70 & 0.76 \\
Mean & Mean & 0.77 & 0.69 & 0.76 \\
Last & Mean & 0.76 & 0.73 & 0.75 \\
Last & Max & 0.74 & 0.73 & 0.74 \\
Mean & Max & 0.74 & 0.68 & 0.73 \\
Max & Max & 0.73 & 0.69 & 0.73 \\
Last & Last & 0.71 & 0.69 & 0.70 \\
Mean & Last & 0.71 & 0.68 & 0.70 \\
Max & Last & 0.70 & 0.67 & 0.70 \\
\bottomrule
\end{tabular}

            \end{minipage}
        }%
        \hfill %
        \subcaptionbox{Average Accuracy Rank (Overall benchmark)}{%
            \includegraphics[width=0.58\textwidth]{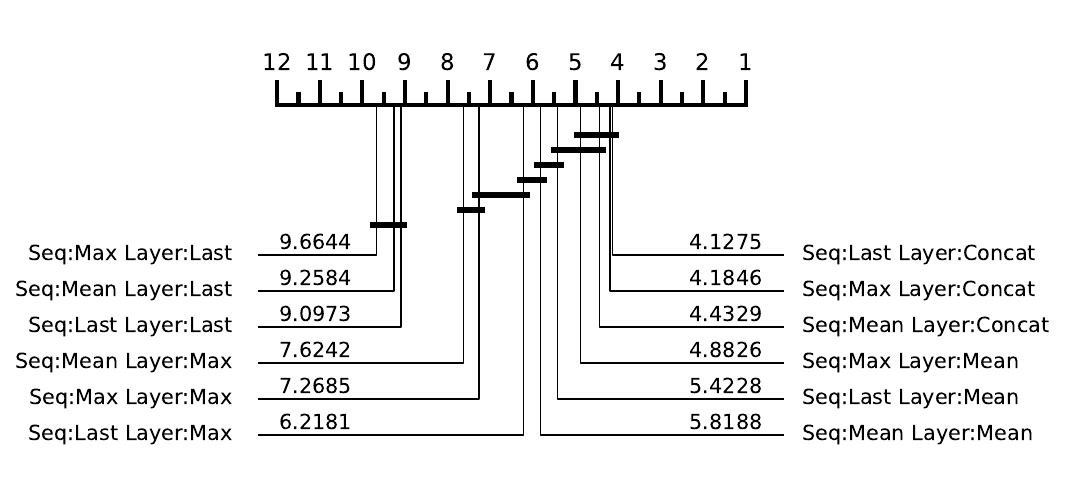}
        }
        \caption{\aggcaption{TimesFM 2.0}}
        \label{fig:ablation-agg-TimesFM2}
    \end{figure}

    \begin{figure}[htbp]
        \centering
        \subcaptionbox{Average Accuracy}{%
            \begin{minipage}{0.4\textwidth}
                \centering
                \small
                \begin{tabular}{llrrr}
\toprule
 &  & Uni & Multi & Comb \\
Seq & Layer &  &  &  \\
\midrule
Mean & Concat & \textbf{0.74} & 0.71 & \textbf{0.73} \\
Last & Concat & 0.73 & \textbf{0.72} & \textbf{0.73} \\
Max & Concat & 0.73 & 0.69 & \textbf{0.73} \\
Mean & Mean & 0.73 & 0.70 & 0.72 \\
Max & Mean & 0.73 & 0.69 & 0.72 \\
Last & Mean & 0.72 & 0.71 & 0.72 \\
Mean & Max & 0.72 & 0.68 & 0.71 \\
Last & Max & 0.71 & 0.69 & 0.71 \\
Max & Max & 0.71 & 0.66 & 0.71 \\
Mean & Last & 0.69 & 0.68 & 0.69 \\
Last & Last & 0.69 & 0.67 & 0.69 \\
Max & Last & 0.68 & 0.66 & 0.68 \\
\bottomrule
\end{tabular}

            \end{minipage}
        }%
        \hfill %
        \subcaptionbox{Average Accuracy Rank (Overall benchmark)}{%
            \includegraphics[width=0.58\textwidth]{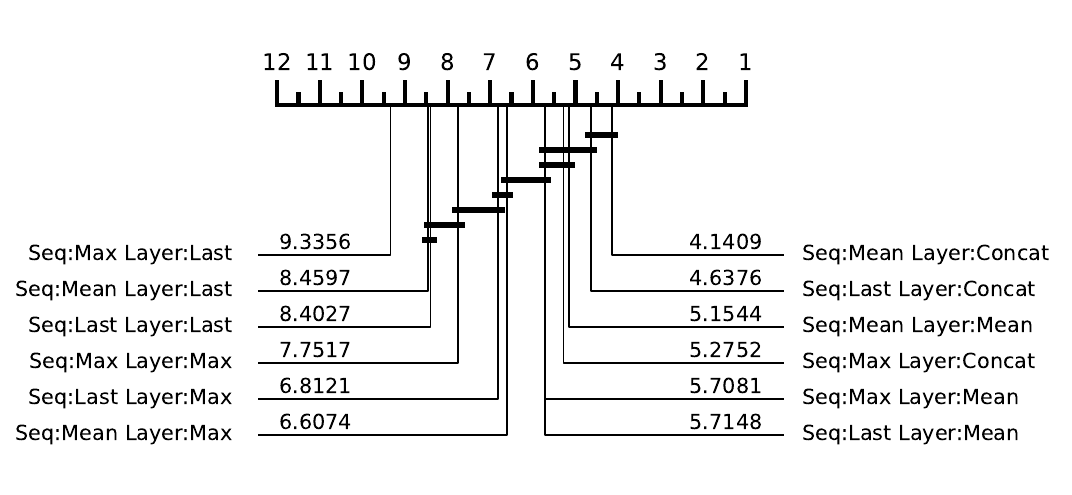}
        }
        \caption{\aggcaption{TimesFM 1.0}}
        \label{fig:ablation-agg-TimesFM1}
    \end{figure}

    \begin{figure}[htbp]
        \centering
        \subcaptionbox{Average Accuracy}{%
            \begin{minipage}{0.4\textwidth}
                \centering
                \small
                \begin{tabular}{llrrr}
\toprule
 &  & Uni & Multi & Comb \\
Seq & Layer &  &  &  \\
\midrule
Mean & Concat & \textbf{0.71} & 0.71 & \textbf{0.71} \\
Mean & Mean & 0.70 & 0.70 & 0.70 \\
Max & Concat & 0.69 & \textbf{0.72} & 0.69 \\
Last & Concat & 0.69 & \textbf{0.72} & 0.69 \\
Mean & Max & 0.68 & 0.70 & 0.68 \\
Max & Mean & 0.67 & 0.71 & 0.68 \\
Last & Mean & 0.67 & 0.71 & 0.67 \\
Mean & Last & 0.66 & 0.68 & 0.66 \\
Last & Max & 0.65 & 0.71 & 0.66 \\
Max & Max & 0.64 & 0.68 & 0.64 \\
Last & Last & 0.62 & 0.70 & 0.64 \\
Max & Last & 0.62 & 0.67 & 0.63 \\
\bottomrule
\end{tabular}

            \end{minipage}
        }%
        \hfill %
        \subcaptionbox{Average Accuracy Rank (Overall benchmark)}{%
            \includegraphics[width=0.58\textwidth]{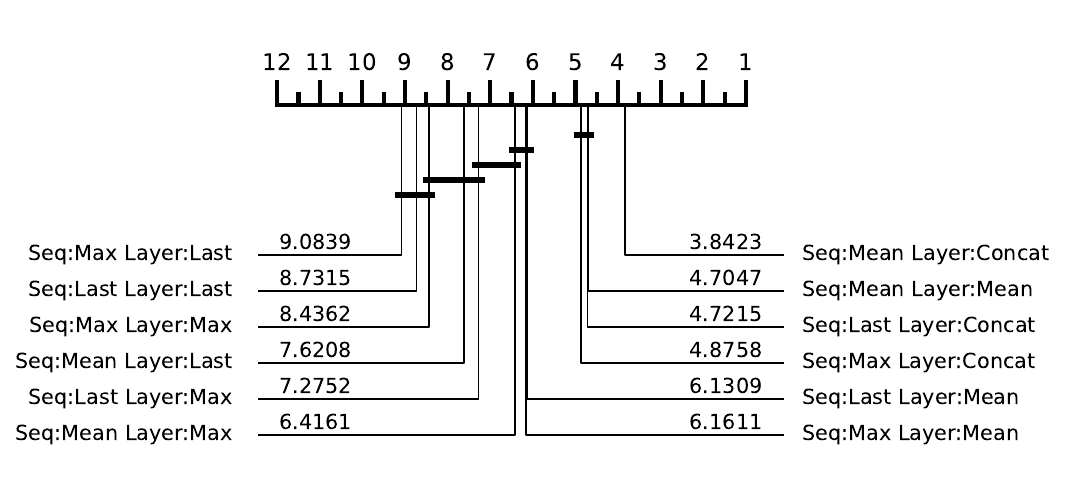}
        }
        \caption{\aggcaption{ToTo}}
        \label{fig:ablation-agg-toto}
    \end{figure}

    \begin{figure}[htbp]
        \centering
        \subcaptionbox{Average Accuracy}{%
            \begin{minipage}{0.4\textwidth}
                \centering
                \small
                \begin{tabular}{lrrr}
\toprule
 & Uni & Multi & Comb \\
Seq &  &  &  \\
\midrule
Mean & \textbf{0.77} & \textbf{0.72} & \textbf{0.76} \\
Max & 0.76 & 0.71 & 0.75 \\
Last & 0.74 & \textbf{0.72} & 0.74 \\
\bottomrule
\end{tabular}

            \end{minipage}
        }%
        \hfill %
        \subcaptionbox{Average Accuracy Rank (Overall benchmark)}{%
            \includegraphics[width=0.58\textwidth]{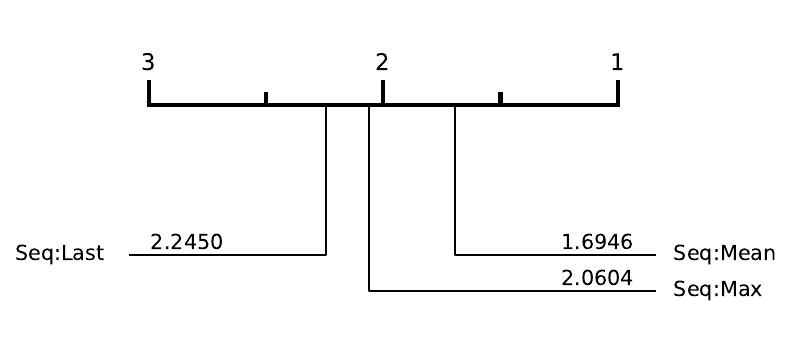}
        }
        \caption{\aggcaption{Chronos Bolt (Base)}}
        \label{fig:ablation-agg-boltbase}
    \end{figure}

    \begin{figure}[htbp]
        \centering
        \subcaptionbox{Average Accuracy}{%
            \begin{minipage}{0.4\textwidth}
                \centering
                \small
                \begin{tabular}{lrrr}
\toprule
 & Uni & Multi & Comb \\
Seq &  &  &  \\
\midrule
Mean & \textbf{0.77} & \textbf{0.73} & \textbf{0.76} \\
Max & 0.76 & \textbf{0.73} & \textbf{0.76} \\
Last & 0.72 & 0.71 & 0.72 \\
\bottomrule
\end{tabular}

            \end{minipage}
        }%
        \hfill %
        \subcaptionbox{Average Accuracy Rank (Overall benchmark)}{%
            \includegraphics[width=0.58\textwidth]{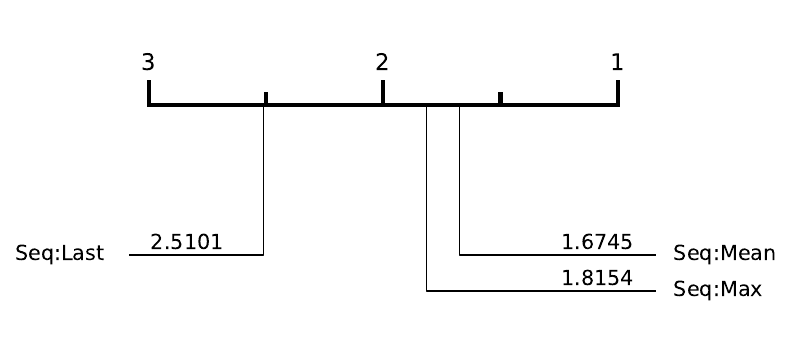}
        }
        \caption{\aggcaption{Chronos Bolt (Small)}}
        \label{fig:ablation-agg-boltsmall}
    \end{figure}

    \begin{figure}[htbp]
        \centering
        \subcaptionbox{Average Accuracy}{%
            \begin{minipage}{0.4\textwidth}
                \centering
                \small
                \begin{tabular}{lrrr}
\toprule
 & Uni & Multi & Comb \\
Seq &  &  &  \\
\midrule
Mean & \textbf{0.71} & \textbf{0.71} & \textbf{0.71} \\
Max & 0.68 & 0.68 & 0.68 \\
Last & 0.59 & 0.65 & 0.60 \\
\bottomrule
\end{tabular}

            \end{minipage}
        }%
        \hfill %
        \subcaptionbox{Average Accuracy Rank (Overall benchmark)}{%
            \includegraphics[width=0.58\textwidth]{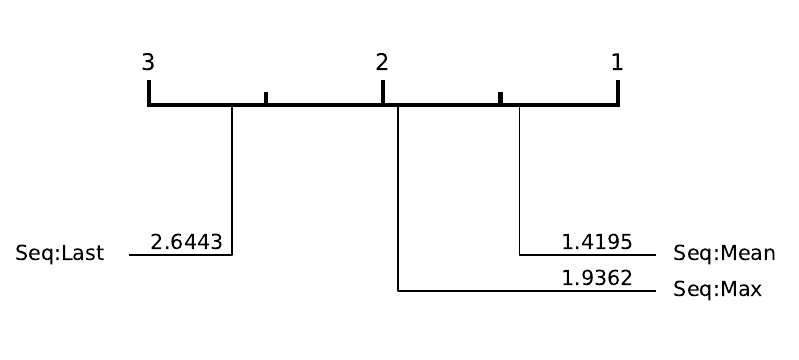}
        }
        \caption{\aggcaption{Chronos (Base)}}
        \label{fig:ablation-agg-chronosbase}
    \end{figure}

    \begin{figure}[htbp]
        \centering
        \subcaptionbox{Average Accuracy}{%
            \begin{minipage}{0.4\textwidth}
                \centering
                \small
                \begin{tabular}{lrrr}
\toprule
 & Uni & Multi & Comb \\
Seq &  &  &  \\
\midrule
Mean & \textbf{0.70} & \textbf{0.70} & \textbf{0.70} \\
Max & 0.67 & 0.67 & 0.67 \\
Last & 0.60 & 0.64 & 0.61 \\
\bottomrule
\end{tabular}

            \end{minipage}
        }%
        \hfill %
        \subcaptionbox{Average Accuracy Rank (Overall benchmark)}{%
            \includegraphics[width=0.58\textwidth]{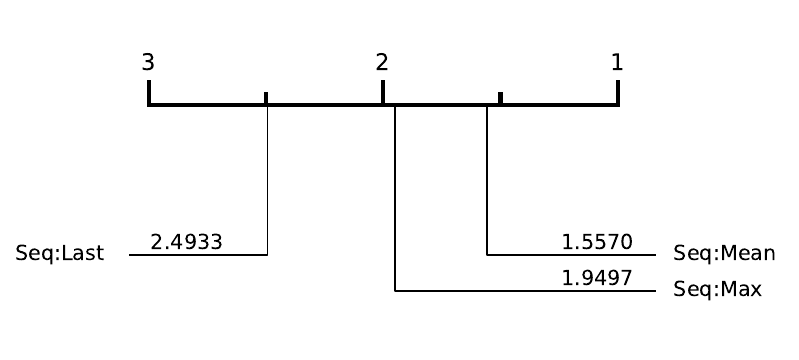}
        }
        \caption{\aggcaption{Chronos (Small)}}
        \label{fig:ablation-agg-chronossmall}
    \end{figure}

\clearpage

\paragraph{Variate aggregation methods}
We conduct an ablation study of the method to aggregate  per-variate embeddings into a single feature vector for a multivariate time series. 
This is necessary when applying a univariate model to each variate independently or when a multivariate model produces distinct per-variate outputs.
We evaluate three strategies: mean pooling, max pooling, and concatenation.
Figures~\ref{fig:var-agg-tirex}-\ref{fig:var-agg-Toto} present the results.
Each figure presents a table with the mean accuracy over univariate, multivariate, and all datasets, complemented by a critical difference plot of mean ranks to visualize statistical significance.
Concatenation consistently outperforms both pooling methods across all tested models.

\newcommand{\varaggcaption}[1]{%
    Results for \textbf{#1} for the \textbf{variate aggregation ablation} experiments. 
    (a) Average accuracy on univariate (Uni), multivariate (Multi), and overall (Comb) benchmark datasets. Sorted by overall accuracy.
    (b) Critical difference diagram of the average accuracy ranks.%
}

    \begin{figure}[htbp]
        \centering
        \subcaptionbox{Average Accuracy}{%
            \begin{minipage}{0.4\textwidth}
                \centering
                \small
                \begin{tabular}{lr}
\toprule
 & Multi \\
Var &  \\
\midrule
Concat & \textbf{0.74} \\
Mean & 0.67 \\
Max & 0.66 \\
\bottomrule
\end{tabular}

            \end{minipage}
        }%
        \hfill %
        \subcaptionbox{Average Accuracy Rank (Multivariate Benchmark)}{%
            \includegraphics[width=0.58\textwidth]{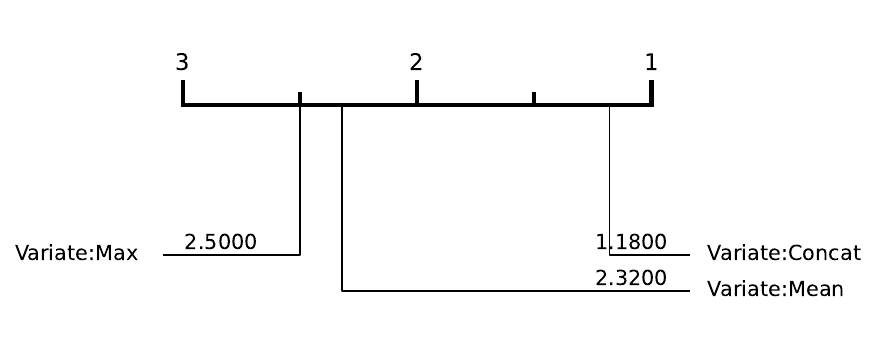}
        }
        \caption{\varaggcaption{TiRex}}
        \label{fig:var-agg-tirex}
    \end{figure}

    \begin{figure}[htbp]
        \centering
        \subcaptionbox{Average Accuracy}{%
            \begin{minipage}{0.4\textwidth}
                \centering
                \small
                \begin{tabular}{lr}
\toprule
 & Multi \\
Var &  \\
\midrule
Concat & \textbf{0.72} \\
Mean & 0.66 \\
Max & 0.65 \\
\bottomrule
\end{tabular}

            \end{minipage}
        }%
        \hfill %
        \subcaptionbox{Average Accuracy Rank (Multivariate Benchmark)}{%
            \includegraphics[width=0.58\textwidth]{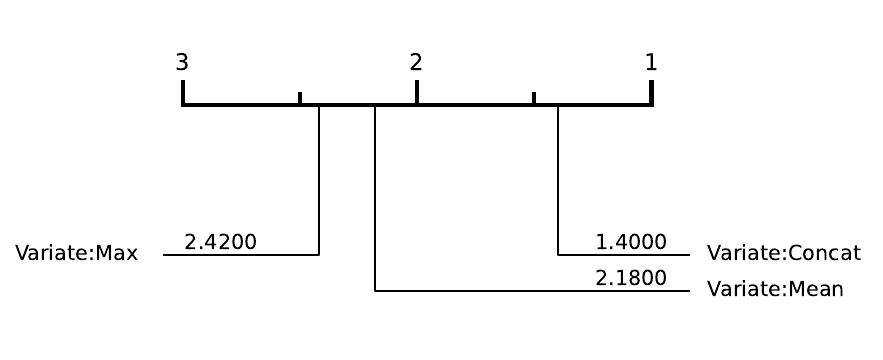}
        }
        \caption{\varaggcaption{Chronos Bolt (Base)}}
        \label{fig:var-agg-ChronosBoltBase}
    \end{figure}

    \begin{figure}[htbp]
        \centering
        \subcaptionbox{Average Accuracy}{%
            \begin{minipage}{0.4\textwidth}
                \centering
                \small
                \begin{tabular}{lr}
\toprule
 & Multi \\
Var &  \\
\midrule
Concat & \textbf{0.73} \\
Max & 0.66 \\
Mean & 0.65 \\
\bottomrule
\end{tabular}

            \end{minipage}
        }%
        \hfill %
        \subcaptionbox{Average Accuracy Rank (Multivariate Benchmark)}{%
            \includegraphics[width=0.58\textwidth]{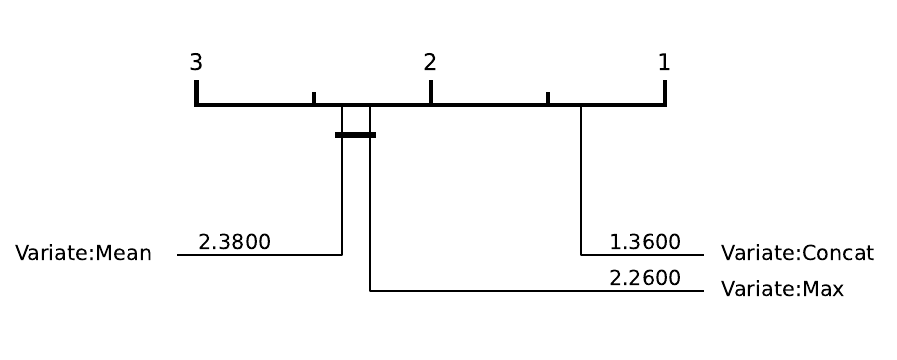}
        }
        \caption{\varaggcaption{Chronos Bolt (Small)}}
        \label{fig:var-agg-ChronosBoltSmall}
    \end{figure}

    \begin{figure}[htbp]
        \centering
        \subcaptionbox{Average Accuracy}{%
            \begin{minipage}{0.4\textwidth}
                \centering
                \small
                \begin{tabular}{lr}
\toprule
 & Multi \\
Var &  \\
\midrule
Concat & \textbf{0.71} \\
Mean & 0.65 \\
Max & 0.65 \\
\bottomrule
\end{tabular}

            \end{minipage}
        }%
        \hfill %
        \subcaptionbox{Average Accuracy Rank (Multivariate Benchmark)}{%
            \includegraphics[width=0.58\textwidth]{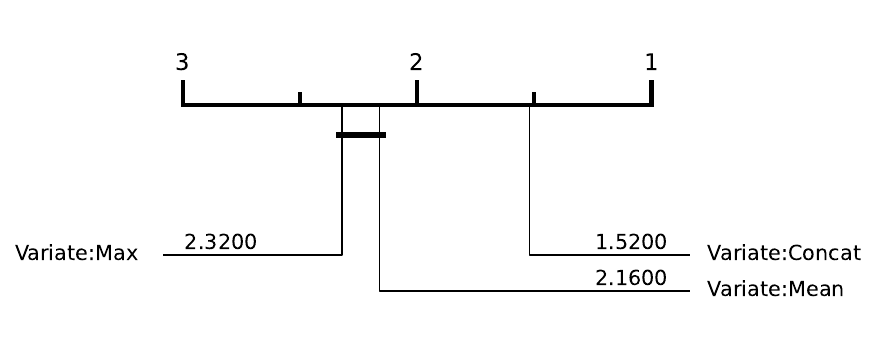}
        }
        \caption{\varaggcaption{Chronos (Base)}}
        \label{fig:var-agg-ChronosBase}
    \end{figure}

    \begin{figure}[htbp]
        \centering
        \subcaptionbox{Average Accuracy}{%
            \begin{minipage}{0.4\textwidth}
                \centering
                \small
                \begin{tabular}{lr}
\toprule
 & Multi \\
Var &  \\
\midrule
Concat & \textbf{0.70} \\
Mean & 0.64 \\
Max & 0.63 \\
\bottomrule
\end{tabular}

            \end{minipage}
        }%
        \hfill %
        \subcaptionbox{Average Accuracy Rank (Multivariate Benchmark)}{%
            \includegraphics[width=0.58\textwidth]{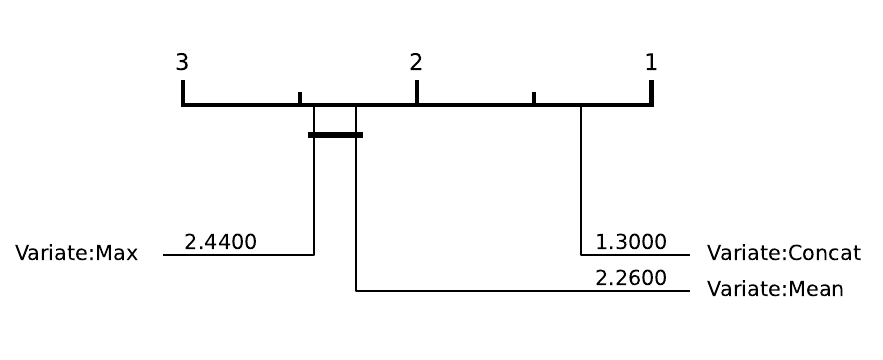}
        }
        \caption{\varaggcaption{Chronos (Small)}}
        \label{fig:var-agg-ChronosSmall}
    \end{figure}

    \begin{figure}[htbp]
        \centering
        \subcaptionbox{Average Accuracy}{%
            \begin{minipage}{0.4\textwidth}
                \centering
                \small
                \begin{tabular}{lr}
\toprule
 & Multi \\
Var &  \\
\midrule
Concat & \textbf{0.71} \\
Mean & 0.65 \\
Max & 0.64 \\
\bottomrule
\end{tabular}

            \end{minipage}
        }%
        \hfill %
        \subcaptionbox{Average Accuracy Rank (Multivariate Benchmark)}{%
            \includegraphics[width=0.58\textwidth]{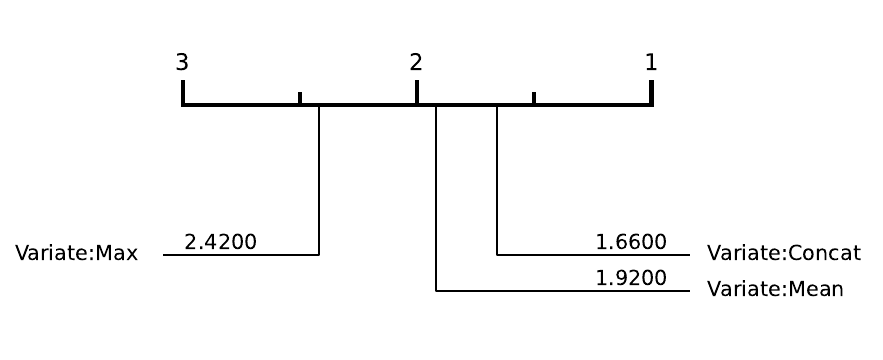}
        }
        \caption{\varaggcaption{TimesFM 1.0}}
        \label{fig:var-agg-timesfm1}
    \end{figure}

    \begin{figure}[htbp]
        \centering
        \subcaptionbox{Average Accuracy}{%
            \begin{minipage}{0.4\textwidth}
                \centering
                \small
                \begin{tabular}{lr}
\toprule
 & Multi \\
Var &  \\
\midrule
Concat & \textbf{0.70} \\
Max & 0.67 \\
Mean & 0.66 \\
\bottomrule
\end{tabular}

            \end{minipage}
        }%
        \hfill %
        \subcaptionbox{Average Accuracy Rank (Multivariate Benchmark)}{%
            \includegraphics[width=0.58\textwidth]{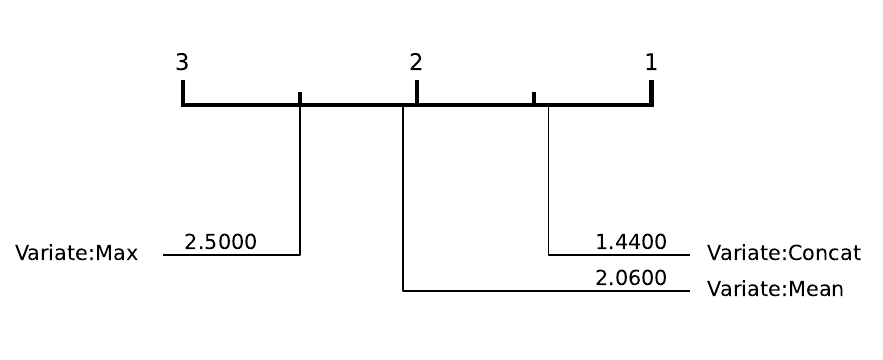}
        }
        \caption{\varaggcaption{TimesFM 2.0}}
        \label{fig:var-agg-timesfm2}
    \end{figure}

    \begin{figure}[htbp]
        \centering
        \subcaptionbox{Average Accuracy}{%
            \begin{minipage}{0.4\textwidth}
                \centering
                \small
                \begin{tabular}{lr}
\toprule
 & Multi \\
Var &  \\
\midrule
Concat & \textbf{0.71} \\
Mean & 0.68 \\
Max & 0.66 \\
\bottomrule
\end{tabular}

            \end{minipage}
        }%
        \hfill %
        \subcaptionbox{Average Accuracy Rank (Multivariate Benchmark)}{%
            \includegraphics[width=0.58\textwidth]{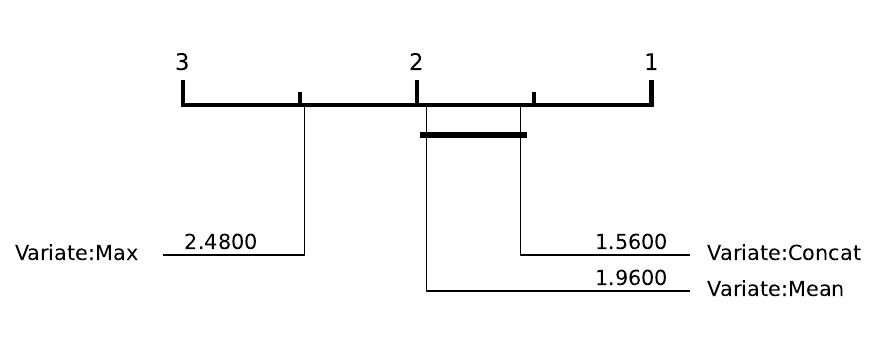}
        }
        \caption{\varaggcaption{ToTo}}
        \label{fig:var-agg-Toto}
    \end{figure}

\clearpage

\subsection{Full Results: Embedding Augmentation}\label{app:ext-results-augment}
In this section, we provide an ablation study of our proposed embedding augmentations.
First, the impact augmentations, both individually and combined, are analyzed quantitatively for each model.
Then we provide a hyperparameter ablation for the Absolute Sample Statistics Augmentation and a qualitative analysis of its impact.

\paragraph{Ablation on individual models} 
We conduct an ablation study to evaluate the effectiveness of our two proposed embedding augmentations.
Figures~\ref{fig:ablation-ext-tirex}-\ref{fig:ablation-ext-toto} present the results.
Each figure presents a table with the mean accuracy over univariate, multivariate, and all datasets, complemented by a critical difference plot of mean ranks to visualize statistical significance.
Both the statistics-based and the differencing-based augmentations individually improve performance for a majority of the models, although the statistical significance of these gains varies.
The combination of both augmentations most often yields further improvements, resulting in the best overall performance.

\newcommand{\extablationcaption}[1]{%
    Results for \textbf{#1} for the \textbf{embedding augmentation ablation} experiments.
    \emph{Diff} and \emph{Stats} indicate the application of the ``differencing'' and the ``sample statistics'' augmentations respectively. 
    (a) Average accuracy on univariate (Uni), multivariate (Multi), and overall (Comb) benchmark datasets. Sorted by overall accuracy. 
    (b) Critical difference diagram of the average accuracy ranks.%
}

    \begin{figure}[htbp]
        \centering
        \subcaptionbox{Average Accuracy}{%
            \begin{minipage}{0.4\textwidth}
                \centering
                \small
                \begin{tabular}{llrrr}
\toprule
 &  & Uni & Multi & Comb \\
Diff & Stats &  &  &  \\
\midrule
True & True & \textbf{0.81} & \textbf{0.74} & \textbf{0.80} \\
True & False & \textbf{0.81} & 0.73 & 0.79 \\
False & True & 0.80 & 0.73 & 0.79 \\
False & False & 0.80 & \textbf{0.74} & 0.79 \\
\bottomrule
\end{tabular}

            \end{minipage}
        }%
        \hfill %
        \subcaptionbox{Average Accuracy Rank (Overall benchmark)}{%
            \includegraphics[width=0.58\textwidth]{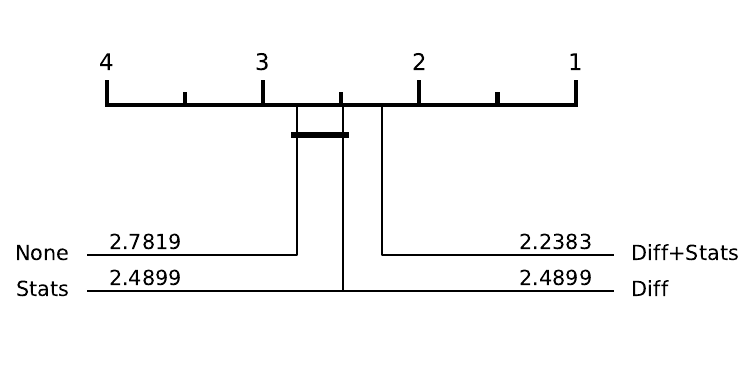}
        }
        \caption{\extablationcaption{TiRex}}
        \label{fig:ablation-ext-tirex}
    \end{figure}

    \begin{figure}[htbp]
        \centering
        \subcaptionbox{Average Accuracy}{%
            \begin{minipage}{0.4\textwidth}
                \centering
                \small
                \begin{tabular}{llrrr}
\toprule
 &  & Uni & Multi & Comb \\
Diff & Stats &  &  &  \\
\midrule
True & True & \textbf{0.79} & \textbf{0.74} & \textbf{0.78} \\
True & False & \textbf{0.79} & 0.72 & 0.77 \\
False & True & 0.78 & \textbf{0.74} & 0.77 \\
False & False & 0.77 & 0.72 & 0.76 \\
\bottomrule
\end{tabular}

            \end{minipage}
        }%
        \hfill %
        \subcaptionbox{Average Accuracy Rank (Overall benchmark)}{%
            \includegraphics[width=0.58\textwidth]{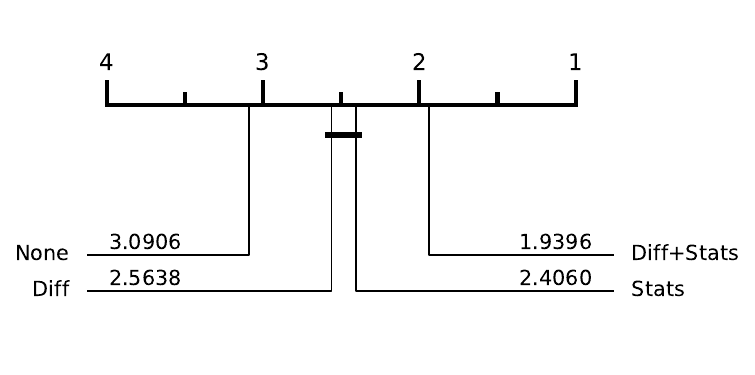}
        }
        \caption{\extablationcaption{Chronos Bolt (Base)}}
        \label{fig:ablation-ext-boltbase}
    \end{figure}

    \begin{figure}[htbp]
        \centering
        \subcaptionbox{Average Accuracy}{%
            \begin{minipage}{0.4\textwidth}
                \centering
                \small
                \begin{tabular}{llrrr}
\toprule
 &  & Uni & Multi & Comb \\
Diff & Stats &  &  &  \\
\midrule
True & True & \textbf{0.79} & \textbf{0.74} & \textbf{0.78} \\
False & True & 0.78 & \textbf{0.74} & 0.77 \\
True & False & 0.78 & 0.72 & 0.77 \\
False & False & 0.77 & 0.73 & 0.76 \\
\bottomrule
\end{tabular}

            \end{minipage}
        }%
        \hfill %
        \subcaptionbox{Average Accuracy Rank (Overall benchmark)}{%
            \includegraphics[width=0.58\textwidth]{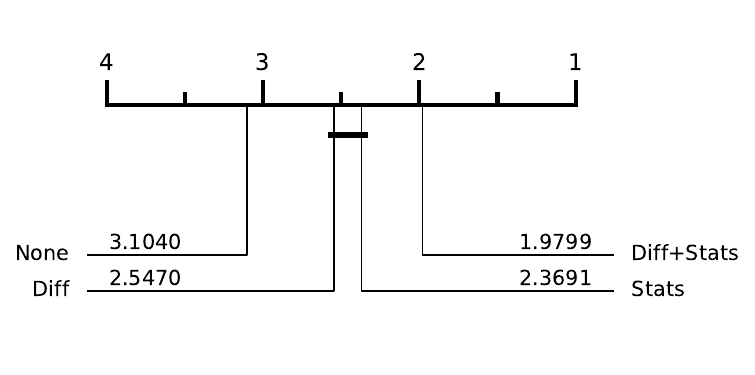}
        }
        \caption{\extablationcaption{Chronos Bolt (Small)}}
        \label{fig:ablation-ext-boltsmall}
    \end{figure}

    \begin{figure}[htbp]
        \centering
        \subcaptionbox{Average Accuracy}{%
            \begin{minipage}{0.4\textwidth}
                \centering
                \small
                \begin{tabular}{llrrr}
\toprule
 &  & Uni & Multi & Comb \\
Diff & Stats &  &  &  \\
\midrule
True & True & \textbf{0.76} & 0.72 & \textbf{0.75} \\
False & True & 0.75 & \textbf{0.73} & \textbf{0.75} \\
True & False & 0.74 & 0.71 & 0.73 \\
False & False & 0.71 & 0.71 & 0.71 \\
\bottomrule
\end{tabular}

            \end{minipage}
        }%
        \hfill %
        \subcaptionbox{Average Accuracy Rank (Overall benchmark)}{%
            \includegraphics[width=0.58\textwidth]{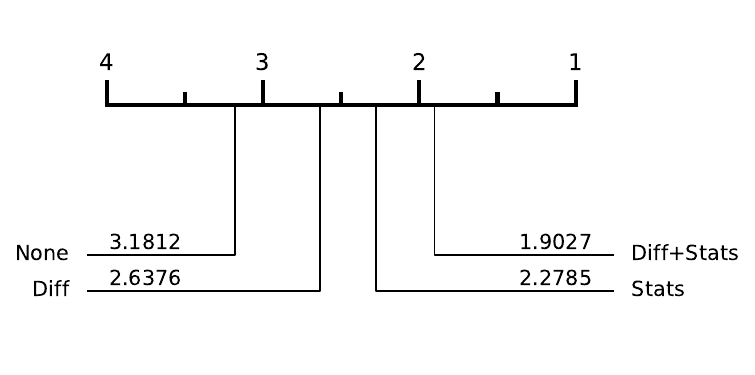}
        }
        \caption{\extablationcaption{Chronos (Base)}}
        \label{fig:ablation-ext-chronosbase}
    \end{figure}

    \begin{figure}[htbp]
        \centering
        \subcaptionbox{Average Accuracy}{%
            \begin{minipage}{0.4\textwidth}
                \centering
                \small
                \begin{tabular}{llrrr}
\toprule
 &  & Uni & Multi & Comb \\
Diff & Stats &  &  &  \\
\midrule
True & True & \textbf{0.75} & 0.72 & \textbf{0.75} \\
False & True & 0.74 & \textbf{0.73} & 0.74 \\
True & False & 0.72 & 0.71 & 0.72 \\
False & False & 0.70 & 0.70 & 0.70 \\
\bottomrule
\end{tabular}

            \end{minipage}
        }%
        \hfill %
        \subcaptionbox{Average Accuracy Rank (Overall benchmark)}{%
            \includegraphics[width=0.58\textwidth]{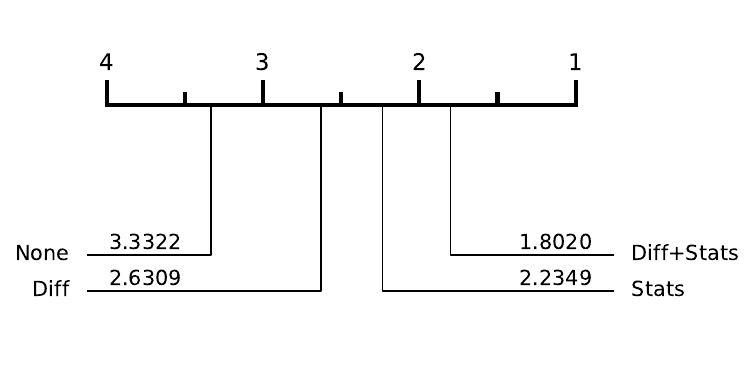}
        }
        \caption{\extablationcaption{Chronos (Small)}}
        \label{fig:ablation-ext-chronossmall}
    \end{figure}

    \begin{figure}[htbp]
        \centering
        \subcaptionbox{Average Accuracy}{%
            \begin{minipage}{0.4\textwidth}
                \centering
                \small
                \begin{tabular}{llrrr}
\toprule
 &  & Uni & Multi & Comb \\
Diff & Stats &  &  &  \\
\midrule
True & True & \textbf{0.79} & 0.70 & \textbf{0.78} \\
False & True & \textbf{0.79} & \textbf{0.71} & \textbf{0.78} \\
True & False & \textbf{0.79} & 0.70 & \textbf{0.78} \\
False & False & \textbf{0.79} & 0.70 & 0.77 \\
\bottomrule
\end{tabular}

            \end{minipage}
        }%
        \hfill %
        \subcaptionbox{Average Accuracy Rank (Overall benchmark)}{%
            \includegraphics[width=0.58\textwidth]{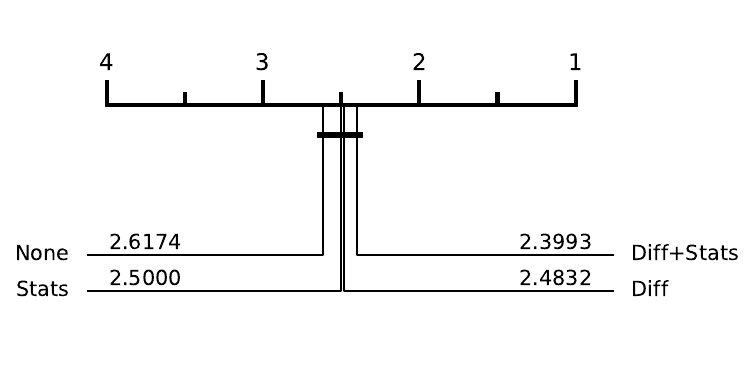}
        }
        \caption{\extablationcaption{TimesFM 2.0}}
        \label{fig:ablation-ext-timesfm2}
    \end{figure}

    \begin{figure}[htbp]
        \centering
        \subcaptionbox{Average Accuracy}{%
            \begin{minipage}{0.4\textwidth}
                \centering
                \small
                \begin{tabular}{llrrr}
\toprule
 &  & Uni & Multi & Comb \\
Diff & Stats &  &  &  \\
\midrule
True & True & \textbf{0.75} & \textbf{0.72} & \textbf{0.74} \\
False & True & \textbf{0.75} & 0.70 & \textbf{0.74} \\
True & False & \textbf{0.75} & 0.70 & \textbf{0.74} \\
False & False & 0.74 & 0.71 & 0.73 \\
\bottomrule
\end{tabular}

            \end{minipage}
        }%
        \hfill %
        \subcaptionbox{Average Accuracy Rank (Overall benchmark)}{%
            \includegraphics[width=0.58\textwidth]{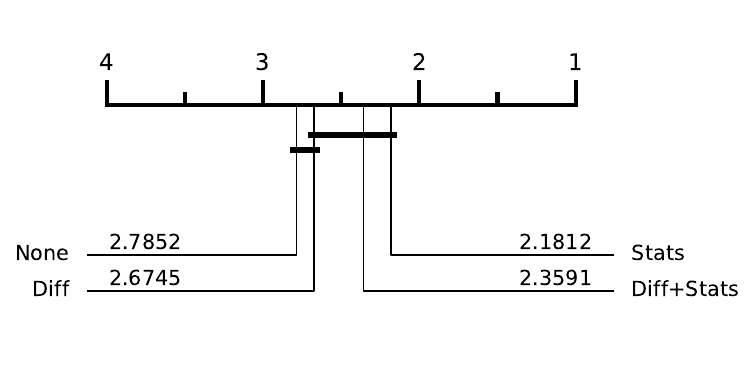}
        }
        \caption{\extablationcaption{TimesFM 1.0}}
        \label{fig:ablation-ext-timesfm1}
    \end{figure}

    \begin{figure}[htbp]
        \centering
        \subcaptionbox{Average Accuracy}{%
            \begin{minipage}{0.4\textwidth}
                \centering
                \small
                \begin{tabular}{llrrr}
\toprule
 &  & Uni & Multi & Comb \\
Diff & Stats &  &  &  \\
\midrule
True & True & \textbf{0.79} & 0.71 & \textbf{0.78} \\
False & True & \textbf{0.79} & \textbf{0.72} & \textbf{0.78} \\
True & False & \textbf{0.79} & 0.70 & \textbf{0.78} \\
False & False & \textbf{0.79} & 0.69 & 0.77 \\
\bottomrule
\end{tabular}

            \end{minipage}
        }%
        \hfill %
        \subcaptionbox{Average Accuracy Rank (Overall benchmark)}{%
            \includegraphics[width=0.58\textwidth]{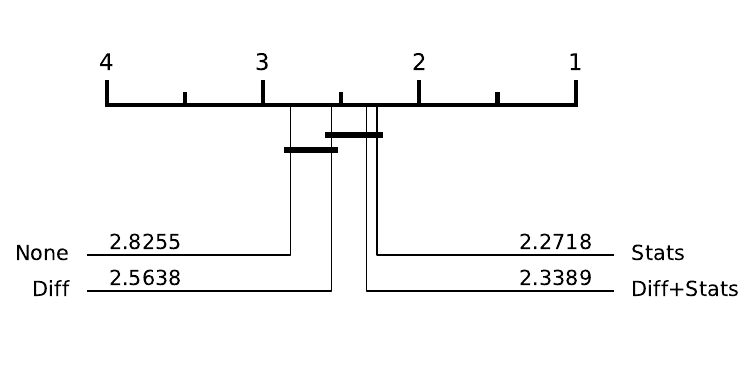}
        }
        \caption{\extablationcaption{Moirai 1.1 (Base)}}
        \label{fig:ablation-ext-moiraibase11}
    \end{figure}

    \begin{figure}[htbp]
        \centering
        \subcaptionbox{Average Accuracy}{%
            \begin{minipage}{0.4\textwidth}
                \centering
                \small
                \begin{tabular}{llrrr}
\toprule
 &  & Uni & Multi & Comb \\
Diff & Stats &  &  &  \\
\midrule
True & True & \textbf{0.74} & 0.70 & \textbf{0.73} \\
True & False & 0.73 & \textbf{0.71} & \textbf{0.73} \\
False & True & 0.72 & \textbf{0.71} & 0.72 \\
False & False & 0.71 & \textbf{0.71} & 0.71 \\
\bottomrule
\end{tabular}

            \end{minipage}
        }%
        \hfill %
        \subcaptionbox{Average Accuracy Rank (Overall benchmark)}{%
            \includegraphics[width=0.58\textwidth]{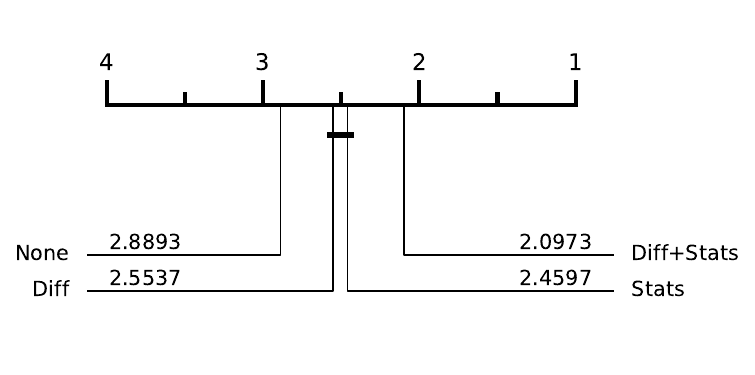}
        }
        \caption{\extablationcaption{ToTo}}
        \label{fig:ablation-ext-toto}
    \end{figure}

\clearpage

\paragraph{Absolute Sample Statistics Augmentation: Number of Patches}

\begin{figure}[htbp]
        \centering
        \subcaptionbox{Average Accuracy}{%
            \begin{minipage}{0.4\textwidth}
                \centering
                \small
                \begin{tabular}{lrrr}
\toprule
$k$ & Uni & Multi & Comb \\
\midrule
32 & \textbf{0.81} & 0.73 & \textbf{0.80} \\
16 & \textbf{0.81} & \textbf{0.74} & \textbf{0.80} \\
4 & \textbf{0.81} & 0.73 & \textbf{0.80} \\
8 & \textbf{0.81} & \textbf{0.74} & \textbf{0.80} \\
2 & 0.80 & \textbf{0.74} & 0.79 \\
1 & \textbf{0.81} & 0.72 & 0.79 \\
\bottomrule
\end{tabular}

            \end{minipage}
        }%
        \hfill %
        \subcaptionbox{Average Accuracy Rank (Overall benchmark)}{%
            \includegraphics[width=0.58\textwidth]{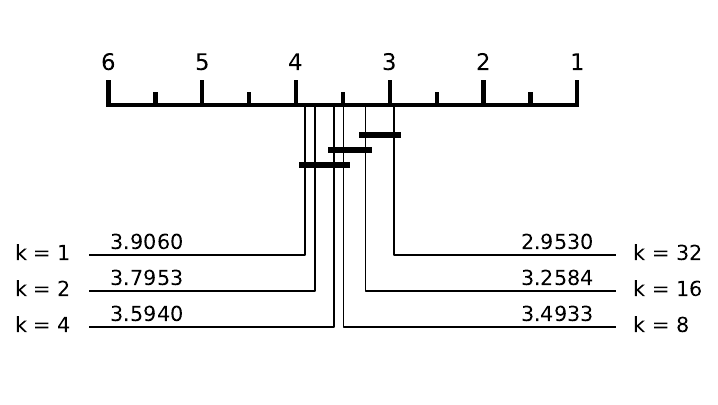}
        }
        \caption{Result of the ablation experiment regarding the number of patches for the absolute sample statistics augmentation
    (a) Average accuracy on univariate (Uni), multivariate (Multi), and overall (Comb) benchmark datasets. Sorted by overall accuracy.
    (b) Critical difference diagram of the average accuracy ranks.%
    }
        \label{fig:ablation-patch-hyp}
    \end{figure}

The absolute sample statistics augmentation divides each time series into $k$ non-overlapping patches.
For the main experiments, we used a fixed value of $k=8$.
To analyze the impact of this choice, we conducted an ablation study on our best-performing model, TiRex, by evaluating $k~\in \{1,2,4,8,16,32\}$.
The results are presented in Figure~\ref{fig:ablation-patch-hyp}. 
While the average ranks suggest that a higher number of patches could marginally improve performance, the average accuracies remain very similar across all settings.
This indicates that the procedure is generally robust to the choice of $k$, although we note that tuning this hyperparameter for specific datasets could be advantageous in a practical application.

\begin{figure}[htbp]
    \centering

    \includegraphics[width=\linewidth]{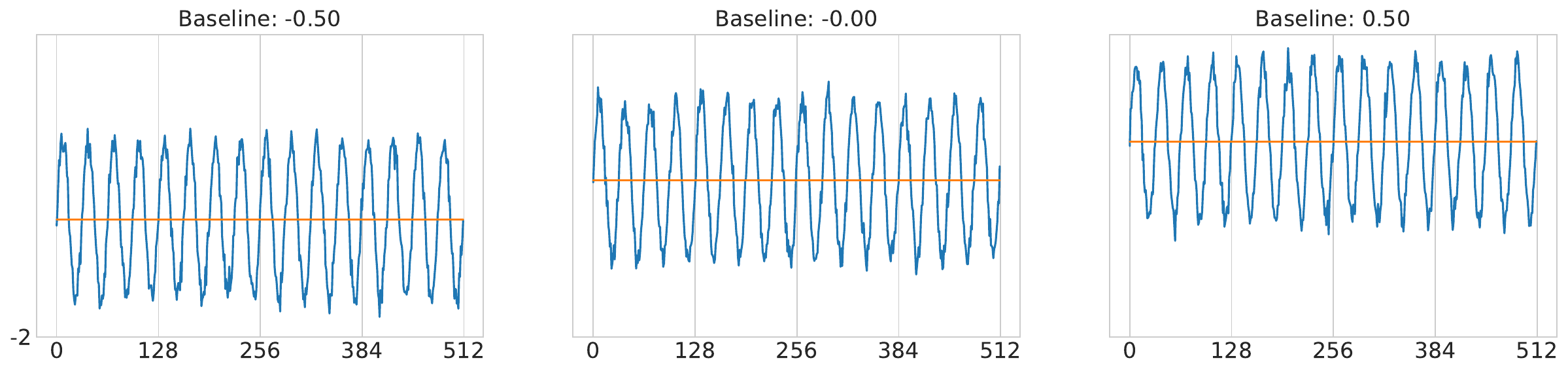}
    \caption{Illustration of three example time series from our synthetic toy dataset. For each series only the baseline value differs between them.}
    \label{fig:baselineshift-examples}

    \vspace{0.75cm} 

    \includegraphics[width=0.95\linewidth]{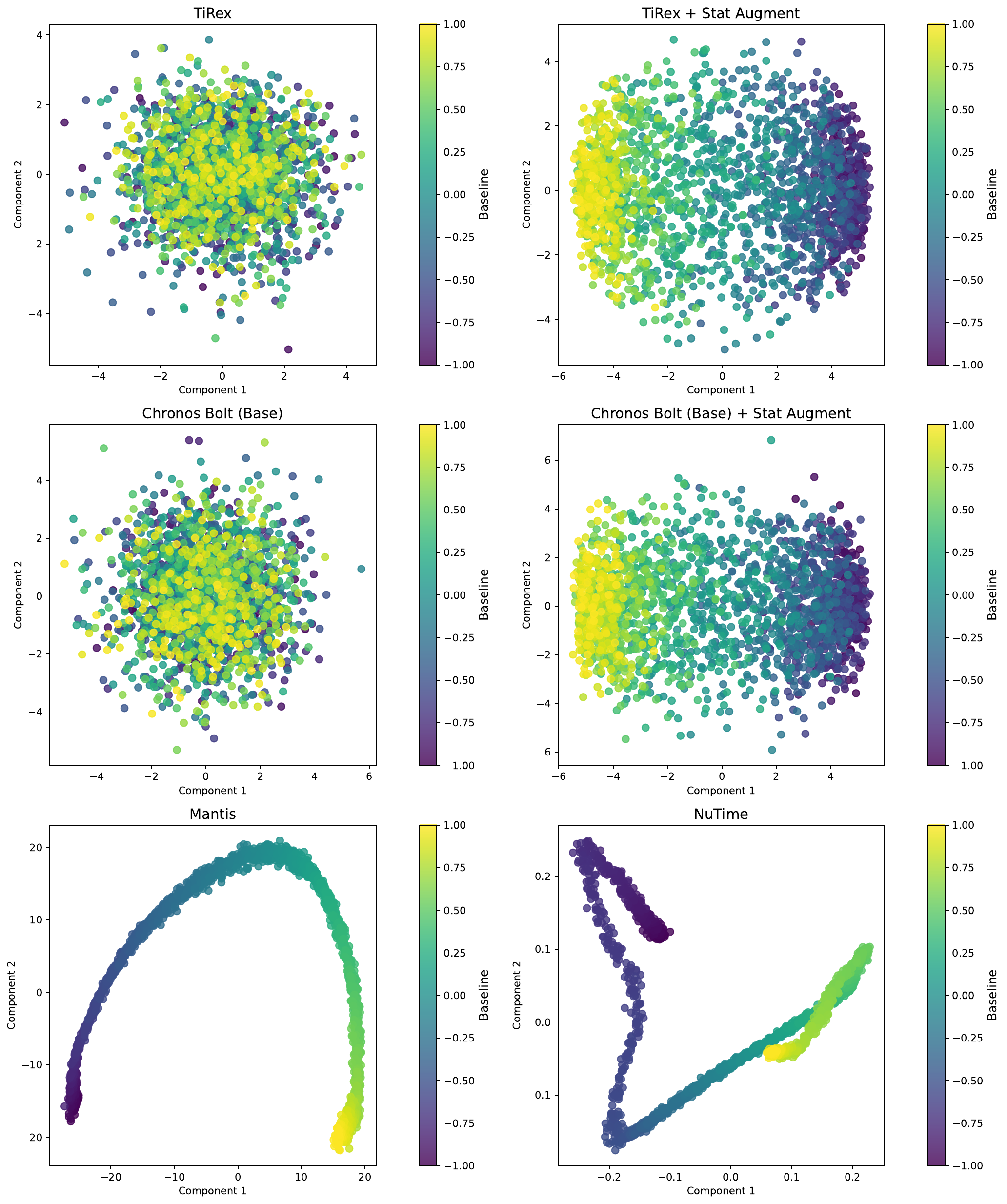}
    \caption{2D PCA projections of embeddings from the baseline-shifted sine wave dataset.
    The left column of the top two row shows the original embeddings from each model, while the right column shows the same embeddings enhanced with our sample statistics augmentation.
    The bottom row shows the embeddings of the pre-trained classification models --- which also allow for separation in terms of this property.}
    \label{fig:baselineshift-pca}
\end{figure}

\paragraph{Absolute Sample Statistics Augmentation: Qualitative Analysis}
As discussed in Section~\ref{sec:method}, instance normalization removes a signal's absolute scale information, such as its mean value. 
To visually demonstrate this effect and the efficacy of our statistics augmentation, we created a toy dataset composed of sine waves that differ only by their baseline value \cite{goswami2024moment}.
Each series is generated using the formula $y_t = sin(5t) + a$, where the baseline a is sampled uniquely for each of the $1024$ examples.
Figure~\ref{fig:baselineshift-examples} shows three such series.

We then generated embeddings for this dataset using TiRex and Chronos Bolt, once without and once with our statistics augmentation, and visualized the results using PCA.
The projections in Figure~\ref{fig:baselineshift-pca} illustrate the outcome.
Without the augmentation, the embeddings from the forecasting models (TiRex, Chronos Bolt) form a single, inseparable cluster.
In contrast, the augmented embeddings show a gradient along the first principal component that directly corresponds to the baseline value $a$.
Notably, the pre-trained classification models also cluster series with similar baselines, i.e., incorporate this property in their representation.

\clearpage

\subsection{Main Results: Balanced Accuracy}\label{app:ext-results-diffmetrcis}

While accuracy is the primary metric in our main evaluation, for consistency with related literature, we also re-evaluated our main experiments using balanced accuracy to ensure the robustness of our findings.
The results are presented in Table~\ref{tab:main-results-balanced}.
The relative performance rankings of the models remain highly consistent across both metrics, with slight changes in the multivariate benchmark data, confirming the robustness of our conclusions.

\begin{table}
    \centering
    \small
    \begin{tabular}{lcccccccc}
\toprule
 & Type & ZS & \multicolumn{2}{c}{Univariate} & \multicolumn{2}{c}{Multivariate} & \multicolumn{2}{c}{Overall} \\
 &  &  & No Aug & Stat+Diff & No Aug & Stat+Diff & No Aug & Stat+Diff \\
\midrule
TiRex & Dec & yes & \textbf{0.78} & \textbf{0.78} & 0.71 & \textbf{0.72} & \textbf{0.77} & \textbf{0.77} \\
Chr. Bolt (Base) & EncDec & yes & 0.75 & 0.77 & 0.70 & \textbf{0.72} & 0.74 & 0.76 \\
Moirai (Large) & Enc & yes & 0.77 & \textbf{0.78} & 0.68 & 0.68 & 0.76 & 0.76 \\
TimesFM 2.0 & Dec & yes & 0.76 & 0.77 & 0.68 & 0.68 & 0.75 & 0.75 \\
TimesFM 1.0 & Dec & yes & 0.72 & 0.72 & 0.68 & 0.69 & 0.71 & 0.72 \\
Chronos (Base) & EncDec & yes & 0.68 & 0.73 & 0.69 & 0.70 & 0.68 & 0.73 \\
ToTo & Dec & yes & 0.68 & 0.71 & 0.69 & 0.69 & 0.68 & 0.70 \\
Mantis & Enc & no &\multicolumn{2}{c}{0.76} & \multicolumn{2}{c}{\textbf{0.72}} & \multicolumn{2}{c}{0.76} \\
NuTime & Enc & no & \multicolumn{2}{c}{0.64} & \multicolumn{2}{c}{0.66} & \multicolumn{2}{c}{0.65} \\
Moment (Large) & Enc & no & \multicolumn{2}{c}{0.59} & \multicolumn{2}{c}{0.55} & \multicolumn{2}{c}{0.58} \\
DTW & - & - & \multicolumn{2}{c}{0.72} &\multicolumn{2}{c}{0.71} & \multicolumn{2}{c}{0.72} \\
\bottomrule
\end{tabular}

    \caption{\textbf{Balanced Accuracy} of different models for the univariate, multivariate, and combined benchmark with a Random Forest Classifier.
    ``Stat+Diff'' shows results with both proposed augmentations applied;  ``no Aug'' utilizes the pure forecasting model representations. ``ZS'' indicates models that did not have access to the benchmark training data during pre-training.}
    \label{tab:main-results-balanced}
\end{table}

\subsection{Main Results: $\leq$ 512 length datasets}\label{app:ext-results-512length}

Several of the evaluated models were pre-trained with a maximum context length of $512$, whereas our full benchmark includes datasets with series up to $2048$ in length.
To assess the impact of this context length discrepancy and to further test the robustness of our findings, we re-ran our main experiments on a subset of the benchmark containing only datasets with a series length of $512$ or less.
The results of this analysis are presented in Table~\ref{tab:main-results-512}.
The relative performance rankings remain consistent with our primary results, with slight changes in the multivariate benchmark data --- this confirms the robustness of our conclusions.

\begin{table}
    \centering
    \small
    \begin{tabular}{lcccccccc}
\toprule
 & Type & ZS & \multicolumn{2}{c}{Univariate} & \multicolumn{2}{c}{Multivariate} & \multicolumn{2}{c}{Overall} \\
 &  &  & No Aug & Stat+Diff & No Aug & Stat+Diff & No Aug & Stat+Diff \\
\midrule
TiRex & Dec & yes & \textbf{0.82} & \textbf{0.83} & 0.77 & 0.77 & \textbf{0.81} & \textbf{0.81} \\
Chr. Bolt (Base) & EncDec & yes & 0.81 & 0.82 & 0.76 & 0.77 & 0.80 & \textbf{0.81} \\
Moirai (Large) & Enc & yes & \textbf{0.82} & 0.82 & 0.74 & 0.74 & 0.80 & \textbf{0.81} \\
TimesFM 2.0 & Dec & yes & 0.81 & 0.81 & 0.74 & 0.73 & 0.80 & 0.80 \\
TimesFM 1.0 & Dec & yes & 0.79 & 0.79 & 0.75 & 0.75 & 0.78 & 0.79 \\
Chronos (Base) & EncDec & yes & 0.74 & 0.79 & 0.74 & 0.76 & 0.74 & 0.78 \\
ToTo & Dec & yes & 0.73 & 0.76 & 0.74 & 0.74 & 0.73 & 0.75 \\
\noalign{\vskip 1mm}\cdashline{1-1}\noalign{\vskip 1mm}
Mantis & Enc & no & \multicolumn{2}{c}{0.81} & \multicolumn{2}{c}{\textbf{0.78}} & \multicolumn{2}{c}{ \textbf{0.81}} \\
NuTime & Enc & no & \multicolumn{2}{c}{0.71} & \multicolumn{2}{c}{0.71} & \multicolumn{2}{c}{0.71} \\
Moment (Large) & Enc & no & \multicolumn{2}{c}{0.68} & \multicolumn{2}{c}{0.60} & \multicolumn{2}{c}{0.66} \\
DTW & - & - & \multicolumn{2}{c}{0.76} & \multicolumn{2}{c}{0.76} & \multicolumn{2}{c}{0.76} \\
\bottomrule
\end{tabular}

    \caption{Classification Accuracy of different models for the univariate, multivariate, and combined benchmark with a Random Forest Classifier --- \textbf{on the subset of datasets with a maximum length of 512}.
    ``Stat+Diff'' shows results with both proposed augmentations applied;  ``no Aug'' utilizes the pure forecasting model representations. ``ZS'' indicates models that did not have access to the benchmark training data during pre-training.}
    \label{tab:main-results-512}
\end{table}

\clearpage

\begin{table}
    \small
    \centering
    \begin{tabular}{lrrrrrrrrr}
\toprule
 & \rotatebox{90}{TiRex} & \rotatebox{90}{Chr. Bolt (Base)} & \rotatebox{90}{Chr. Bolt (Small)} & \rotatebox{90}{Moirai (Large)} & \rotatebox{90}{Moirai (Base)} & \rotatebox{90}{Moirai (Small)} & \rotatebox{90}{TimesFM 2.0} & \rotatebox{90}{TimesFM 1.0} & \rotatebox{90}{Chronos (Base)} \\
\midrule
ACSF1 & 0.85 & 0.82 & 0.82 & 0.88 & 0.86 & 0.86 & 0.83 & 0.75 & 0.84 \\
Adiac & 0.78 & 0.79 & 0.80 & 0.79 & 0.79 & 0.78 & 0.78 & 0.79 & 0.71 \\
ArrowHead & 0.78 & 0.83 & 0.81 & 0.78 & 0.77 & 0.76 & 0.73 & 0.74 & 0.66 \\
Beef & 0.80 & 0.67 & 0.73 & 0.70 & 0.67 & 0.60 & 0.80 & 0.83 & 0.60 \\
BeetleFly & 0.90 & 0.90 & 0.90 & 0.95 & 0.95 & 0.85 & 0.95 & 0.85 & 0.75 \\
BirdChicken & 0.90 & 0.90 & 0.95 & 0.90 & 0.90 & 0.90 & 0.80 & 0.90 & 0.90 \\
BME & 0.99 & 1.00 & 1.00 & 0.99 & 0.96 & 0.95 & 0.95 & 0.95 & 0.98 \\
Car & 0.80 & 0.82 & 0.77 & 0.78 & 0.75 & 0.68 & 0.82 & 0.78 & 0.83 \\
CBF & 0.99 & 1.00 & 0.97 & 1.00 & 1.00 & 0.96 & 1.00 & 0.99 & 0.96 \\
Chinatown & 0.97 & 0.98 & 0.99 & 0.96 & 0.97 & 0.95 & 0.97 & 0.97 & 0.97 \\
ChlorineConcentration & 0.72 & 0.71 & 0.72 & 0.74 & 0.75 & 0.75 & 0.69 & 0.69 & 0.64 \\
CinCECGTorso & 0.99 & 0.85 & 0.90 & 0.84 & 0.83 & 0.75 & 0.98 & 0.96 & 0.97 \\
Coffee & 1.00 & 0.96 & 1.00 & 1.00 & 0.96 & 0.96 & 0.96 & 1.00 & 0.96 \\
Computers & 0.76 & 0.72 & 0.73 & 0.77 & 0.76 & 0.77 & 0.72 & 0.70 & 0.74 \\
CricketX & 0.71 & 0.71 & 0.69 & 0.69 & 0.68 & 0.62 & 0.69 & 0.64 & 0.62 \\
CricketY & 0.74 & 0.73 & 0.70 & 0.69 & 0.67 & 0.61 & 0.75 & 0.69 & 0.69 \\
CricketZ & 0.72 & 0.73 & 0.72 & 0.74 & 0.69 & 0.69 & 0.69 & 0.64 & 0.63 \\
Crop & 0.74 & 0.74 & 0.74 & NaN & 0.73 & 0.73 & 0.73 & NaN & 0.71 \\
DiatomSizeReduction & 0.86 & 0.90 & 0.92 & 0.87 & 0.82 & 0.85 & 0.81 & 0.85 & 0.88 \\
DistalPhalanxOutlineCorrect & 0.79 & 0.77 & 0.80 & 0.80 & 0.80 & 0.78 & 0.79 & 0.78 & 0.76 \\
DistalPhalanxOutlineAgeGroup & 0.76 & 0.76 & 0.75 & 0.78 & 0.74 & 0.71 & 0.75 & 0.76 & 0.76 \\
DistalPhalanxTW & 0.65 & 0.66 & 0.68 & 0.68 & 0.72 & 0.69 & 0.65 & 0.66 & 0.66 \\
Earthquakes & 0.75 & 0.76 & 0.76 & 0.72 & 0.74 & 0.75 & 0.74 & 0.76 & 0.74 \\
ECG200 & 0.85 & 0.85 & 0.85 & 0.84 & 0.86 & 0.83 & 0.87 & 0.86 & 0.77 \\
ECG5000 & 0.94 & 0.94 & 0.94 & 0.94 & 0.94 & 0.93 & 0.93 & 0.94 & 0.93 \\
ECGFiveDays & 0.83 & 0.86 & 0.90 & 0.92 & 0.84 & 0.81 & 0.91 & 0.75 & 0.77 \\
ElectricDevices & 0.70 & 0.70 & 0.69 & NaN & 0.72 & 0.70 & 0.70 & 0.66 & 0.74 \\
EOGHorizontalSignal & 0.54 & 0.54 & 0.56 & 0.60 & 0.57 & 0.48 & 0.66 & 0.21 & 0.36 \\
EOGVerticalSignal & 0.42 & 0.43 & 0.45 & 0.42 & 0.44 & 0.40 & 0.46 & 0.15 & 0.30 \\
EthanolLevel & 0.37 & 0.44 & 0.43 & 0.43 & 0.42 & 0.55 & 0.35 & 0.57 & 0.56 \\
FaceAll & 0.86 & 0.71 & 0.71 & 0.72 & 0.78 & 0.70 & 0.83 & 0.85 & 0.69 \\
FaceFour & 0.68 & 0.76 & 0.64 & 0.72 & 0.68 & 0.65 & 0.81 & 0.74 & 0.62 \\
FacesUCR & 0.81 & 0.75 & 0.76 & 0.76 & 0.73 & 0.72 & 0.73 & 0.77 & 0.73 \\
FiftyWords & 0.62 & 0.67 & 0.69 & 0.60 & 0.58 & 0.53 & 0.60 & 0.57 & 0.64 \\
Fish & 0.93 & 0.85 & 0.85 & 0.93 & 0.88 & 0.84 & 0.90 & 0.90 & 0.89 \\
FordA & 0.95 & 0.94 & 0.93 & 0.93 & 0.93 & 0.90 & 0.94 & 0.95 & 0.94 \\
FordB & 0.84 & 0.78 & 0.80 & 0.81 & 0.81 & 0.78 & 0.84 & 0.82 & 0.77 \\
FreezerRegularTrain & 0.93 & 0.92 & 0.91 & 0.97 & 0.97 & 0.91 & 0.88 & 0.85 & 0.97 \\
FreezerSmallTrain & 0.81 & 0.85 & 0.84 & 0.87 & 0.84 & 0.79 & 0.78 & 0.69 & 0.87 \\
GunPoint & 0.95 & 0.94 & 0.99 & 0.97 & 0.97 & 0.95 & 0.95 & 0.88 & 0.95 \\
GunPointAgeSpan & 0.98 & 0.99 & 0.98 & 0.98 & 0.97 & 0.97 & 0.95 & 0.94 & 0.98 \\
GunPointMaleVersusFemale & 1.00 & 1.00 & 1.00 & 0.99 & 1.00 & 0.99 & 1.00 & 0.99 & 0.99 \\
GunPointOldVersusYoung & 0.98 & 1.00 & 1.00 & 0.99 & 0.99 & 0.99 & 0.96 & 0.93 & 0.99 \\
Ham & 0.65 & 0.71 & 0.64 & 0.57 & 0.61 & 0.68 & 0.60 & 0.55 & 0.60 \\
Haptics & 0.52 & 0.51 & 0.51 & 0.51 & 0.57 & 0.49 & 0.54 & 0.51 & 0.51 \\
Herring & 0.59 & 0.67 & 0.67 & 0.59 & 0.59 & 0.62 & 0.61 & 0.53 & 0.59 \\
HouseTwenty & 0.97 & 0.89 & 0.93 & 0.97 & 0.96 & 0.94 & 0.87 & 0.60 & 0.72 \\
InlineSkate & 0.44 & 0.54 & 0.49 & 0.49 & 0.44 & 0.43 & 0.50 & 0.37 & 0.39 \\
InsectEPGRegularTrain & 0.99 & 1.00 & 1.00 & 1.00 & 1.00 & 1.00 & 1.00 & 0.99 & 1.00 \\
InsectEPGSmallTrain & 0.93 & 0.93 & 0.92 & 0.96 & 0.94 & 0.94 & 0.90 & 0.86 & 0.99 \\
\bottomrule
\end{tabular}

    \caption{Accuracy results of the different models with augmentations on the individual datasets with a Random Forest classifier. (Part 1/6)}
    \label{tab:individual-part1}
\end{table}
\begin{table}
    \small
    \centering
    \begin{tabular}{lrrrrrrrrr}
\toprule
 & \rotatebox{90}{Chronos (Small)} & \rotatebox{90}{ToTo} & \rotatebox{90}{Mantis} & \rotatebox{90}{NuTime} & \rotatebox{90}{Moment (Large)} & \rotatebox{90}{Moment (Base)} & \rotatebox{90}{Moment (Large)} & \rotatebox{90}{DTW (1-NN)} & \rotatebox{90}{DTW (3-NN)} \\
\midrule
ACSF1 & 0.84 & 0.79 & 0.79 & 0.78 & 0.43 & 0.55 & 0.48 & 0.64 & 0.59 \\
Adiac & 0.72 & 0.55 & 0.74 & 0.70 & 0.08 & 0.12 & 0.08 & 0.60 & 0.56 \\
ArrowHead & 0.61 & 0.80 & 0.73 & 0.74 & 0.59 & 0.50 & 0.58 & 0.70 & 0.71 \\
Beef & 0.60 & 0.70 & 0.63 & 0.97 & 0.43 & 0.40 & 0.50 & 0.63 & 0.57 \\
BeetleFly & 0.75 & 0.95 & 0.85 & 0.70 & 0.85 & 0.95 & 0.90 & 0.70 & 0.70 \\
BirdChicken & 0.95 & 0.85 & 1.00 & 0.55 & 0.80 & 0.85 & 0.65 & 0.75 & 0.60 \\
BME & 0.98 & 0.97 & 0.92 & 0.93 & 0.79 & 0.81 & 0.88 & 0.89 & 0.85 \\
Car & 0.85 & 0.75 & 0.83 & 0.62 & 0.67 & 0.58 & 0.60 & 0.73 & 0.55 \\
CBF & 0.98 & 0.99 & 0.99 & 0.54 & 0.90 & 0.94 & 0.86 & 1.00 & 1.00 \\
Chinatown & 0.97 & 0.82 & 0.85 & 0.98 & 0.77 & 0.87 & 0.84 & 0.97 & 0.97 \\
ChlorineConcentration & 0.62 & 0.60 & 0.68 & 0.76 & 0.56 & 0.55 & 0.55 & 0.65 & 0.57 \\
CinCECGTorso & 0.95 & 0.90 & 0.67 & 0.58 & 0.57 & 0.68 & 0.68 & 0.65 & 0.50 \\
Coffee & 0.86 & 0.93 & 0.96 & 1.00 & 0.89 & 0.96 & 0.82 & 1.00 & 0.93 \\
Computers & 0.71 & 0.74 & 0.74 & 0.71 & 0.63 & 0.65 & 0.66 & 0.70 & 0.71 \\
CricketX & 0.68 & 0.59 & 0.75 & 0.24 & 0.59 & 0.64 & 0.65 & 0.75 & 0.74 \\
CricketY & 0.68 & 0.63 & 0.73 & 0.36 & 0.61 & 0.60 & 0.56 & 0.74 & 0.70 \\
CricketZ & 0.68 & 0.58 & 0.79 & 0.26 & 0.62 & 0.64 & 0.67 & 0.75 & 0.74 \\
Crop & 0.72 & 0.69 & 0.68 & 0.74 & 0.50 & 0.56 & 0.55 & 0.68 & 0.66 \\
DiatomSizeReduction & 0.87 & 0.83 & 0.87 & 0.87 & 0.50 & 0.50 & 0.56 & 0.97 & 0.93 \\
DistalPhalanxOutlineCorrect & 0.78 & 0.77 & 0.74 & 0.78 & 0.63 & 0.65 & 0.62 & 0.72 & 0.74 \\
DistalPhalanxOutlineAgeGroup & 0.75 & 0.74 & 0.79 & 0.77 & 0.63 & 0.67 & 0.65 & 0.77 & 0.73 \\
DistalPhalanxTW & 0.66 & 0.68 & 0.69 & 0.71 & 0.58 & 0.58 & 0.56 & 0.59 & 0.62 \\
Earthquakes & 0.75 & 0.75 & 0.75 & 0.75 & 0.75 & 0.74 & 0.75 & 0.72 & 0.74 \\
ECG200 & 0.81 & 0.85 & 0.81 & 0.80 & 0.81 & 0.82 & 0.82 & 0.77 & 0.80 \\
ECG5000 & 0.93 & 0.92 & 0.92 & 0.93 & 0.93 & 0.92 & 0.93 & 0.92 & 0.94 \\
ECGFiveDays & 0.82 & 0.60 & 0.93 & 0.76 & 0.65 & 0.88 & 0.74 & 0.77 & 0.62 \\
ElectricDevices & 0.73 & 0.68 & 0.73 & 0.65 & 0.59 & 0.59 & 0.59 & 0.60 & 0.61 \\
EOGHorizontalSignal & 0.45 & 0.49 & 0.58 & 0.33 & 0.07 & 0.10 & 0.11 & 0.44 & 0.43 \\
EOGVerticalSignal & 0.27 & 0.39 & 0.47 & 0.25 & 0.10 & 0.10 & 0.11 & 0.43 & 0.44 \\
EthanolLevel & 0.37 & 0.33 & 0.29 & 0.60 & 0.25 & 0.27 & 0.25 & 0.28 & 0.26 \\
FaceAll & 0.71 & 0.75 & 0.78 & 0.78 & 0.57 & 0.53 & 0.48 & 0.81 & 0.81 \\
FaceFour & 0.56 & 0.57 & 0.95 & 0.62 & 0.65 & 0.55 & 0.57 & 0.83 & 0.68 \\
FacesUCR & 0.82 & 0.63 & 0.83 & 0.67 & 0.54 & 0.48 & 0.47 & 0.90 & 0.88 \\
FiftyWords & 0.64 & 0.52 & 0.64 & 0.57 & 0.48 & 0.48 & 0.44 & 0.69 & 0.66 \\
Fish & 0.85 & 0.80 & 0.94 & 0.78 & 0.49 & 0.55 & 0.42 & 0.82 & 0.79 \\
FordA & 0.93 & 0.92 & 0.86 & 0.81 & 0.88 & 0.90 & 0.89 & 0.55 & 0.58 \\
FordB & 0.76 & 0.82 & 0.74 & 0.62 & 0.73 & 0.77 & 0.72 & 0.62 & 0.62 \\
FreezerRegularTrain & 0.96 & 0.91 & 0.94 & 0.99 & 0.78 & 0.78 & 0.77 & 0.90 & 0.88 \\
FreezerSmallTrain & 0.88 & 0.78 & 0.80 & 0.96 & 0.75 & 0.76 & 0.76 & 0.76 & 0.73 \\
GunPoint & 0.93 & 0.91 & 0.97 & 0.95 & 0.77 & 0.81 & 0.79 & 0.91 & 0.89 \\
GunPointAgeSpan & 0.97 & 0.91 & 0.99 & 0.88 & 0.88 & 0.86 & 0.85 & 0.98 & 0.99 \\
GunPointMaleVersusFemale & 1.00 & 0.96 & 1.00 & 0.97 & 0.94 & 0.95 & 0.96 & 0.98 & 0.97 \\
GunPointOldVersusYoung & 0.99 & 0.89 & 1.00 & 1.00 & 0.86 & 0.88 & 0.84 & 1.00 & 1.00 \\
Ham & 0.66 & 0.78 & 0.70 & 0.73 & 0.70 & 0.65 & 0.63 & 0.47 & 0.51 \\
Haptics & 0.49 & 0.50 & 0.49 & 0.45 & 0.40 & 0.44 & 0.41 & 0.38 & 0.43 \\
Herring & 0.67 & 0.59 & 0.66 & 0.59 & 0.58 & 0.58 & 0.59 & 0.53 & 0.48 \\
HouseTwenty & 0.71 & 0.97 & 0.95 & 0.65 & 0.55 & 0.65 & 0.62 & 0.84 & 0.85 \\
InlineSkate & 0.38 & 0.40 & 0.39 & 0.25 & 0.20 & 0.21 & 0.20 & 0.38 & 0.36 \\
InsectEPGRegularTrain & 1.00 & 1.00 & 1.00 & 0.82 & 0.89 & 0.92 & 0.90 & 1.00 & 1.00 \\
InsectEPGSmallTrain & 0.99 & 1.00 & 1.00 & 0.80 & 0.81 & 0.90 & 0.92 & 1.00 & 1.00 \\
\bottomrule
\end{tabular}

    \caption{Accuracy results of the different models with augmentations on the individual datasets with a Random Forest classifier. (Part 2/6)}
    \label{tab:individual-part2}
\end{table}
\begin{table}
    \small
    \centering
    \begin{tabular}{lrrrrrrrrr}
\toprule
 & \rotatebox{90}{TiRex} & \rotatebox{90}{Chr. Bolt (Base)} & \rotatebox{90}{Chr. Bolt (Small)} & \rotatebox{90}{Moirai (Large)} & \rotatebox{90}{Moirai (Base)} & \rotatebox{90}{Moirai (Small)} & \rotatebox{90}{TimesFM 2.0} & \rotatebox{90}{TimesFM 1.0} & \rotatebox{90}{Chronos (Base)} \\
\midrule
InsectWingbeatSound & 0.66 & 0.62 & 0.64 & 0.61 & 0.61 & 0.60 & 0.62 & 0.63 & 0.55 \\
ItalyPowerDemand & 0.96 & 0.95 & 0.95 & 0.95 & 0.95 & 0.96 & 0.97 & 0.97 & 0.92 \\
LargeKitchenAppliances & 0.79 & 0.75 & 0.72 & 0.79 & 0.83 & 0.75 & 0.77 & 0.67 & 0.76 \\
Lightning2 & 0.75 & 0.75 & 0.72 & 0.75 & 0.70 & 0.67 & 0.66 & 0.69 & 0.70 \\
Lightning7 & 0.70 & 0.71 & 0.77 & 0.63 & 0.64 & 0.63 & 0.58 & 0.67 & 0.67 \\
Mallat & 0.94 & 0.87 & 0.89 & 0.90 & 0.92 & 0.93 & 0.84 & 0.70 & 0.72 \\
Meat & 0.90 & 0.92 & 0.93 & 1.00 & 0.93 & 0.92 & 0.93 & 0.97 & 0.88 \\
MedicalImages & 0.72 & 0.72 & 0.72 & 0.72 & 0.71 & 0.70 & 0.74 & 0.75 & 0.69 \\
MiddlePhalanxOutlineCorrect & 0.85 & 0.84 & 0.83 & 0.85 & 0.86 & 0.84 & 0.86 & 0.87 & 0.81 \\
MiddlePhalanxOutlineAgeGroup & 0.60 & 0.58 & 0.58 & 0.59 & 0.58 & 0.59 & 0.60 & 0.58 & 0.56 \\
MiddlePhalanxTW & 0.55 & 0.58 & 0.53 & 0.54 & 0.55 & 0.56 & 0.53 & 0.54 & 0.55 \\
MixedShapesRegularTrain & 0.97 & 0.95 & 0.96 & 0.97 & 0.97 & 0.95 & 0.97 & 0.94 & 0.96 \\
MixedShapesSmallTrain & 0.94 & 0.93 & 0.93 & 0.94 & 0.96 & 0.93 & 0.95 & 0.91 & 0.92 \\
MoteStrain & 0.91 & 0.90 & 0.91 & 0.91 & 0.88 & 0.82 & 0.91 & 0.85 & 0.93 \\
NonInvasiveFetalECGThorax1 & 0.92 & 0.89 & 0.89 & 0.91 & 0.89 & 0.88 & 0.90 & 0.76 & 0.84 \\
NonInvasiveFetalECGThorax2 & 0.93 & 0.91 & 0.92 & 0.93 & 0.91 & 0.90 & 0.93 & 0.81 & 0.87 \\
OliveOil & 0.87 & 0.87 & 0.87 & 0.83 & 0.90 & 0.90 & 0.90 & 0.90 & 0.83 \\
OSULeaf & 0.96 & 0.90 & 0.92 & 0.95 & 0.95 & 0.88 & 0.96 & 0.84 & 0.93 \\
PhalangesOutlinesCorrect & 0.83 & 0.83 & 0.81 & 0.84 & 0.84 & 0.83 & 0.84 & 0.82 & 0.77 \\
Phoneme & 0.39 & 0.35 & 0.35 & 0.39 & 0.37 & 0.35 & 0.37 & 0.32 & 0.35 \\
PigAirwayPressure & 0.35 & 0.18 & 0.14 & 0.37 & 0.38 & 0.33 & 0.32 & 0.14 & 0.15 \\
PigArtPressure & 0.91 & 0.33 & 0.34 & 0.87 & 0.88 & 0.84 & 0.81 & 0.41 & 0.58 \\
PigCVP & 0.82 & 0.25 & 0.24 & 0.75 & 0.70 & 0.51 & 0.68 & 0.32 & 0.27 \\
Plane & 1.00 & 1.00 & 1.00 & 1.00 & 1.00 & 1.00 & 1.00 & 1.00 & 1.00 \\
PowerCons & 0.89 & 0.88 & 0.90 & 0.93 & 0.91 & 0.90 & 0.90 & 0.91 & 0.93 \\
ProximalPhalanxOutlineCorrect & 0.89 & 0.85 & 0.85 & 0.89 & 0.89 & 0.90 & 0.89 & 0.87 & 0.85 \\
ProximalPhalanxOutlineAgeGroup & 0.86 & 0.85 & 0.87 & 0.87 & 0.86 & 0.84 & 0.87 & 0.86 & 0.85 \\
ProximalPhalanxTW & 0.83 & 0.82 & 0.82 & 0.80 & 0.81 & 0.82 & 0.81 & 0.81 & 0.80 \\
RefrigerationDevices & 0.58 & 0.57 & 0.58 & 0.52 & 0.53 & 0.55 & 0.55 & 0.51 & 0.59 \\
ScreenType & 0.51 & 0.49 & 0.46 & 0.51 & 0.51 & 0.38 & 0.54 & 0.48 & 0.46 \\
SemgHandGenderCh2 & 0.87 & 0.90 & 0.90 & 0.89 & 0.89 & 0.88 & 0.92 & 0.68 & 0.80 \\
SemgHandMovementCh2 & 0.66 & 0.67 & 0.71 & 0.59 & 0.58 & 0.59 & 0.64 & 0.39 & 0.61 \\
SemgHandSubjectCh2 & 0.81 & 0.83 & 0.84 & 0.80 & 0.78 & 0.79 & 0.80 & 0.51 & 0.69 \\
ShapeletSim & 0.96 & 1.00 & 1.00 & 0.97 & 0.98 & 0.85 & 0.96 & 0.96 & 1.00 \\
ShapesAll & 0.86 & 0.81 & 0.83 & 0.85 & 0.85 & 0.82 & 0.85 & 0.81 & 0.83 \\
SmallKitchenAppliances & 0.82 & 0.81 & 0.82 & 0.83 & 0.78 & 0.81 & 0.83 & 0.79 & 0.81 \\
SmoothSubspace & 0.93 & 0.96 & 0.94 & 0.97 & 0.93 & 0.93 & 0.91 & 0.94 & 0.95 \\
SonyAIBORobotSurface1 & 0.88 & 0.80 & 0.82 & 0.73 & 0.71 & 0.64 & 0.90 & 0.84 & 0.53 \\
SonyAIBORobotSurface2 & 0.86 & 0.90 & 0.86 & 0.91 & 0.86 & 0.85 & 0.90 & 0.90 & 0.89 \\
StarLightCurves & 0.98 & 0.97 & 0.98 & NaN & 0.98 & 0.98 & 0.98 & 0.96 & 0.97 \\
Strawberry & 0.96 & 0.95 & 0.95 & 0.95 & 0.95 & 0.95 & 0.96 & 0.96 & 0.92 \\
SwedishLeaf & 0.94 & 0.92 & 0.94 & 0.96 & 0.95 & 0.93 & 0.95 & 0.95 & 0.93 \\
Symbols & 0.96 & 0.95 & 0.98 & 0.99 & 0.98 & 0.97 & 0.95 & 0.94 & 0.87 \\
SyntheticControl & 0.99 & 0.99 & 0.98 & 0.98 & 0.99 & 0.97 & 0.99 & 0.99 & 0.99 \\
ToeSegmentation1 & 0.93 & 0.88 & 0.82 & 0.95 & 0.95 & 0.86 & 0.89 & 0.88 & 0.93 \\
ToeSegmentation2 & 0.92 & 0.90 & 0.88 & 0.86 & 0.87 & 0.86 & 0.87 & 0.88 & 0.88 \\
Trace & 1.00 & 1.00 & 1.00 & 1.00 & 1.00 & 1.00 & 1.00 & 1.00 & 1.00 \\
TwoLeadECG & 0.96 & 0.91 & 0.87 & 0.94 & 0.87 & 0.79 & 1.00 & 0.95 & 0.92 \\
TwoPatterns & 0.97 & 0.94 & 0.92 & 0.97 & 0.93 & 0.84 & 0.96 & 0.94 & 0.89 \\
UMD & 0.94 & 0.96 & 0.94 & 0.96 & 0.97 & 0.85 & 0.90 & 0.89 & 0.90 \\
\bottomrule
\end{tabular}

    \caption{Accuracy results of the different models with augmentations on the individual datasets with a Random Forest classifier. (Part 3/6)}
    \label{tab:individual-part3}
\end{table}
\begin{table}
    \small
    \centering
    \begin{tabular}{lrrrrrrrrr}
\toprule
 & \rotatebox{90}{Chronos (Small)} & \rotatebox{90}{ToTo} & \rotatebox{90}{Mantis} & \rotatebox{90}{NuTime} & \rotatebox{90}{Moment (Large)} & \rotatebox{90}{Moment (Base)} & \rotatebox{90}{Moment (Large)} & \rotatebox{90}{DTW (1-NN)} & \rotatebox{90}{DTW (3-NN)} \\
\midrule
InsectWingbeatSound & 0.56 & 0.61 & 0.51 & 0.62 & 0.55 & 0.52 & 0.48 & 0.36 & 0.35 \\
ItalyPowerDemand & 0.94 & 0.96 & 0.91 & 0.95 & 0.89 & 0.92 & 0.81 & 0.95 & 0.95 \\
LargeKitchenAppliances & 0.75 & 0.72 & 0.79 & 0.52 & 0.71 & 0.80 & 0.79 & 0.79 & 0.80 \\
Lightning2 & 0.72 & 0.59 & 0.80 & 0.66 & 0.66 & 0.69 & 0.66 & 0.87 & 0.87 \\
Lightning7 & 0.66 & 0.58 & 0.77 & 0.42 & 0.60 & 0.64 & 0.62 & 0.73 & 0.71 \\
Mallat & 0.76 & 0.75 & 0.90 & 0.88 & 0.50 & 0.55 & 0.53 & 0.93 & 0.93 \\
Meat & 0.87 & 0.87 & 0.93 & 0.92 & 0.40 & 0.35 & 0.35 & 0.93 & 0.93 \\
MedicalImages & 0.69 & 0.66 & 0.71 & 0.59 & 0.55 & 0.56 & 0.54 & 0.74 & 0.71 \\
MiddlePhalanxOutlineCorrect & 0.81 & 0.77 & 0.80 & 0.81 & 0.57 & 0.57 & 0.57 & 0.70 & 0.73 \\
MiddlePhalanxOutlineAgeGroup & 0.59 & 0.62 & 0.60 & 0.62 & 0.48 & 0.52 & 0.46 & 0.50 & 0.56 \\
MiddlePhalanxTW & 0.56 & 0.56 & 0.54 & 0.57 & 0.46 & 0.51 & 0.49 & 0.51 & 0.51 \\
MixedShapesRegularTrain & 0.95 & 0.96 & 0.94 & 0.89 & 0.78 & 0.82 & 0.80 & 0.84 & 0.83 \\
MixedShapesSmallTrain & 0.92 & 0.94 & 0.90 & 0.80 & 0.70 & 0.77 & 0.75 & 0.78 & 0.75 \\
MoteStrain & 0.88 & 0.88 & 0.92 & 0.86 & 0.85 & 0.85 & 0.79 & 0.83 & 0.81 \\
NonInvasiveFetalECGThorax1 & 0.83 & 0.82 & 0.61 & 0.86 & 0.29 & 0.46 & 0.35 & 0.79 & 0.79 \\
NonInvasiveFetalECGThorax2 & 0.87 & 0.86 & 0.68 & 0.90 & 0.35 & 0.53 & 0.42 & 0.86 & 0.86 \\
OliveOil & 0.83 & 0.47 & 0.93 & 0.90 & 0.37 & 0.40 & 0.43 & 0.83 & 0.87 \\
OSULeaf & 0.92 & 0.82 & 0.86 & 0.51 & 0.72 & 0.74 & 0.69 & 0.59 & 0.58 \\
PhalangesOutlinesCorrect & 0.75 & 0.74 & 0.77 & 0.81 & 0.62 & 0.64 & 0.63 & 0.73 & 0.76 \\
Phoneme & 0.35 & 0.36 & 0.33 & 0.15 & 0.28 & 0.28 & 0.26 & 0.23 & 0.21 \\
PigAirwayPressure & 0.12 & 0.22 & 0.50 & 0.04 & 0.05 & 0.06 & 0.06 & 0.18 & 0.12 \\
PigArtPressure & 0.49 & 0.50 & 0.91 & 0.09 & 0.22 & 0.37 & 0.31 & 0.48 & 0.36 \\
PigCVP & 0.23 & 0.44 & 0.77 & 0.13 & 0.12 & 0.25 & 0.18 & 0.33 & 0.23 \\
Plane & 0.99 & 0.98 & 1.00 & 0.99 & 0.91 & 0.97 & 0.91 & 1.00 & 1.00 \\
PowerCons & 0.94 & 0.94 & 0.91 & 0.88 & 0.82 & 0.85 & 0.77 & 0.92 & 0.86 \\
ProximalPhalanxOutlineCorrect & 0.80 & 0.80 & 0.80 & 0.90 & 0.69 & 0.73 & 0.68 & 0.78 & 0.83 \\
ProximalPhalanxOutlineAgeGroup & 0.86 & 0.86 & 0.85 & 0.86 & 0.80 & 0.80 & 0.80 & 0.80 & 0.81 \\
ProximalPhalanxTW & 0.80 & 0.81 & 0.78 & 0.80 & 0.60 & 0.67 & 0.59 & 0.76 & 0.77 \\
RefrigerationDevices & 0.53 & 0.58 & 0.51 & 0.46 & 0.51 & 0.56 & 0.54 & 0.46 & 0.46 \\
ScreenType & 0.47 & 0.43 & 0.44 & 0.43 & 0.39 & 0.47 & 0.47 & 0.40 & 0.39 \\
SemgHandGenderCh2 & 0.82 & 0.83 & 0.90 & 0.78 & 0.66 & 0.67 & 0.68 & 0.92 & 0.91 \\
SemgHandMovementCh2 & 0.61 & 0.52 & 0.73 & 0.38 & 0.24 & 0.27 & 0.32 & 0.78 & 0.76 \\
SemgHandSubjectCh2 & 0.72 & 0.72 & 0.79 & 0.56 & 0.38 & 0.34 & 0.32 & 0.87 & 0.85 \\
ShapeletSim & 1.00 & 0.86 & 0.94 & 0.54 & 0.84 & 0.91 & 0.74 & 0.65 & 0.63 \\
ShapesAll & 0.84 & 0.76 & 0.83 & 0.71 & 0.68 & 0.68 & 0.64 & 0.77 & 0.71 \\
SmallKitchenAppliances & 0.82 & 0.79 & 0.81 & 0.78 & 0.70 & 0.72 & 0.74 & 0.64 & 0.67 \\
SmoothSubspace & 0.96 & 0.93 & 0.91 & 0.98 & 0.67 & 0.81 & 0.71 & 0.83 & 0.85 \\
SonyAIBORobotSurface1 & 0.55 & 0.66 & 0.78 & 0.59 & 0.50 & 0.57 & 0.55 & 0.73 & 0.62 \\
SonyAIBORobotSurface2 & 0.82 & 0.78 & 0.87 & 0.82 & 0.83 & 0.84 & 0.84 & 0.83 & 0.80 \\
StarLightCurves & 0.97 & 0.98 & 0.98 & 0.97 & 0.89 & 0.90 & 0.88 & 0.91 & 0.91 \\
Strawberry & 0.93 & 0.91 & 0.95 & 0.95 & 0.71 & 0.77 & 0.67 & 0.94 & 0.92 \\
SwedishLeaf & 0.92 & 0.87 & 0.92 & 0.90 & 0.67 & 0.70 & 0.65 & 0.79 & 0.77 \\
Symbols & 0.87 & 0.91 & 0.97 & 0.85 & 0.88 & 0.95 & 0.91 & 0.95 & 0.93 \\
SyntheticControl & 0.99 & 0.97 & 0.98 & 0.83 & 0.96 & 0.89 & 0.87 & 0.99 & 0.98 \\
ToeSegmentation1 & 0.83 & 0.78 & 0.97 & 0.60 & 0.90 & 0.93 & 0.93 & 0.77 & 0.75 \\
ToeSegmentation2 & 0.65 & 0.87 & 0.95 & 0.58 & 0.88 & 0.85 & 0.88 & 0.84 & 0.82 \\
Trace & 0.99 & 0.93 & 1.00 & 0.51 & 0.89 & 0.99 & 0.96 & 1.00 & 1.00 \\
TwoLeadECG & 0.98 & 0.79 & 1.00 & 0.69 & 0.63 & 0.70 & 0.69 & 0.90 & 0.85 \\
TwoPatterns & 0.80 & 0.87 & 0.88 & 0.57 & 0.86 & 0.83 & 0.76 & 1.00 & 1.00 \\
UMD & 0.81 & 0.91 & 0.97 & 0.81 & 0.83 & 0.88 & 0.85 & 0.88 & 0.85 \\
\bottomrule
\end{tabular}

    \caption{Accuracy results of the different models with augmentations on the individual datasets with a Random Forest classifier. (Part 4/6)}
    \label{tab:individual-part4}
\end{table}
\begin{table}
    \small
    \centering
    \begin{tabular}{lrrrrrrrrr}
\toprule
 & \rotatebox{90}{TiRex} & \rotatebox{90}{Chr. Bolt (Base)} & \rotatebox{90}{Chr. Bolt (Small)} & \rotatebox{90}{Moirai (Large)} & \rotatebox{90}{Moirai (Base)} & \rotatebox{90}{Moirai (Small)} & \rotatebox{90}{TimesFM 2.0} & \rotatebox{90}{TimesFM 1.0} & \rotatebox{90}{Chronos (Base)} \\
\midrule
UWaveGestureLibraryAll & 0.91 & 0.95 & 0.95 & 0.85 & 0.84 & 0.84 & 0.88 & 0.79 & 0.89 \\
UWaveGestureLibraryX & 0.81 & 0.80 & 0.82 & 0.79 & 0.78 & 0.77 & 0.73 & 0.70 & 0.81 \\
UWaveGestureLibraryY & 0.75 & 0.74 & 0.77 & 0.73 & 0.72 & 0.72 & 0.66 & 0.64 & 0.76 \\
UWaveGestureLibraryZ & 0.74 & 0.75 & 0.75 & 0.74 & 0.74 & 0.71 & 0.67 & 0.64 & 0.74 \\
Wafer & 1.00 & 1.00 & 0.99 & 0.99 & 0.99 & 0.99 & 1.00 & 1.00 & 1.00 \\
Wine & 0.72 & 0.78 & 0.61 & 0.72 & 0.70 & 0.81 & 0.85 & 0.76 & 0.54 \\
WordSynonyms & 0.54 & 0.53 & 0.58 & 0.48 & 0.47 & 0.45 & 0.49 & 0.49 & 0.52 \\
Worms & 0.82 & 0.68 & 0.70 & 0.81 & 0.83 & 0.75 & 0.77 & 0.69 & 0.69 \\
WormsTwoClass & 0.84 & 0.82 & 0.81 & 0.81 & 0.81 & 0.83 & 0.82 & 0.75 & 0.78 \\
Yoga & 0.80 & 0.84 & 0.85 & 0.83 & 0.77 & 0.80 & 0.80 & 0.77 & 0.82 \\
AllGestureWiimoteX & 0.60 & 0.62 & 0.61 & 0.62 & 0.60 & 0.54 & 0.67 & 0.64 & 0.53 \\
AllGestureWiimoteY & 0.70 & 0.66 & 0.70 & 0.66 & 0.65 & 0.62 & 0.72 & 0.68 & 0.57 \\
AllGestureWiimoteZ & 0.60 & 0.61 & 0.62 & 0.62 & 0.59 & 0.56 & 0.65 & 0.63 & 0.51 \\
GestureMidAirD1 & 0.74 & 0.81 & 0.72 & 0.75 & 0.71 & 0.68 & 0.61 & 0.49 & 0.62 \\
GestureMidAirD2 & 0.68 & 0.71 & 0.69 & 0.72 & 0.67 & 0.65 & 0.49 & 0.44 & 0.63 \\
GestureMidAirD3 & 0.52 & 0.50 & 0.51 & 0.50 & 0.48 & 0.43 & 0.34 & 0.28 & 0.41 \\
GesturePebbleZ1 & 0.86 & 0.86 & 0.86 & 0.84 & 0.87 & 0.87 & 0.85 & 0.83 & 0.76 \\
GesturePebbleZ2 & 0.86 & 0.82 & 0.82 & 0.87 & 0.85 & 0.90 & 0.80 & 0.82 & 0.71 \\
PickupGestureWiimoteZ & 0.78 & 0.76 & 0.74 & 0.82 & 0.70 & 0.68 & 0.68 & 0.62 & 0.80 \\
ShakeGestureWiimoteZ & 0.86 & 0.88 & 0.86 & 0.84 & 0.82 & 0.82 & 0.92 & 0.90 & 0.88 \\
DodgerLoopDay & 0.47 & 0.52 & 0.57 & 0.49 & 0.43 & 0.49 & 0.48 & 0.42 & 0.48 \\
DodgerLoopGame & 0.69 & 0.84 & 0.76 & 0.75 & 0.80 & 0.70 & 0.62 & 0.65 & 0.70 \\
DodgerLoopWeekend & 0.90 & 0.92 & 0.94 & 0.88 & 0.94 & 0.90 & 0.98 & 0.97 & 0.97 \\
MelbournePedestrian & 0.92 & 0.93 & 0.93 & 0.90 & 0.90 & 0.91 & 0.90 & 0.90 & 0.93 \\
ArticularyWordRecognition & 0.99 & 1.00 & 1.00 & 0.97 & 0.99 & 0.98 & 0.98 & 0.98 & 0.97 \\
AtrialFibrillation & 0.27 & 0.33 & 0.33 & 0.07 & 0.33 & 0.33 & 0.07 & 0.47 & 0.20 \\
BasicMotions & 1.00 & 1.00 & 1.00 & 0.97 & 0.97 & 1.00 & 1.00 & 0.97 & 1.00 \\
Cricket & 1.00 & 0.99 & 1.00 & 0.96 & 0.94 & 0.94 & 1.00 & 0.69 & 0.93 \\
Epilepsy & 1.00 & 1.00 & 1.00 & 0.97 & 0.97 & 1.00 & 0.99 & 0.99 & 0.99 \\
EthanolConcentration & 0.37 & 0.37 & 0.37 & 0.36 & 0.33 & 0.47 & 0.37 & 0.54 & 0.54 \\
ERing & 0.97 & 0.96 & 0.99 & 0.88 & 0.93 & 0.93 & 0.97 & 0.93 & 0.95 \\
FaceDetection & 0.63 & 0.57 & 0.58 & 0.57 & 0.59 & 0.58 & NaN & NaN & 0.57 \\
FingerMovements & 0.48 & 0.54 & 0.52 & 0.56 & 0.48 & 0.51 & 0.52 & 0.53 & 0.47 \\
HandMovementDirection & 0.32 & 0.30 & 0.26 & 0.19 & 0.26 & 0.32 & 0.22 & 0.22 & 0.31 \\
Handwriting & 0.22 & 0.26 & 0.24 & 0.22 & 0.20 & 0.22 & 0.26 & 0.29 & 0.18 \\
Heartbeat & 0.73 & 0.74 & 0.76 & 0.74 & 0.73 & 0.75 & 0.73 & 0.73 & 0.73 \\
Libras & 0.87 & 0.88 & 0.88 & 0.73 & 0.73 & 0.81 & 0.84 & 0.84 & 0.87 \\
LSST & 0.59 & 0.61 & 0.61 & 0.60 & 0.60 & 0.60 & 0.58 & 0.56 & 0.58 \\
NATOPS & 0.81 & 0.82 & 0.89 & 0.86 & 0.82 & 0.84 & 0.80 & 0.83 & 0.82 \\
PenDigits & 0.97 & 0.96 & 0.96 & NaN & 0.95 & 0.95 & 0.97 & 0.96 & 0.95 \\
PEMS-SF & 1.00 & 0.95 & 0.99 & 0.99 & 1.00 & 0.99 & NaN & 1.00 & 0.98 \\
PhonemeSpectra & 0.26 & 0.25 & 0.24 & 0.27 & 0.25 & 0.22 & 0.27 & 0.24 & 0.26 \\
RacketSports & 0.83 & 0.86 & 0.86 & 0.80 & 0.81 & 0.75 & 0.84 & 0.90 & 0.84 \\
SelfRegulationSCP1 & 0.84 & 0.79 & 0.82 & 0.75 & 0.77 & 0.77 & 0.82 & 0.76 & 0.75 \\
SelfRegulationSCP2 & 0.57 & 0.54 & 0.53 & 0.56 & 0.54 & 0.49 & 0.50 & 0.47 & 0.43 \\
UWaveGestureLibrary & 0.88 & 0.86 & 0.86 & 0.81 & 0.83 & 0.88 & 0.68 & 0.68 & 0.81 \\
CharacterTrajectories & 0.96 & 0.97 & 0.98 & 0.93 & 0.93 & 0.96 & 0.95 & 0.95 & 0.96 \\
JapaneseVowels & 0.89 & 0.92 & 0.94 & 0.85 & 0.88 & 0.93 & 0.83 & 0.85 & 0.93 \\
SpokenArabicDigits & 0.96 & 0.97 & 0.97 & NaN & 0.96 & 0.94 & NaN & 0.96 & 0.97 \\
\bottomrule
\end{tabular}

    \caption{Accuracy results of the different models with augmentations on the individual datasets with a Random Forest classifier. (Part 5/6)}
    \label{tab:individual-part5}
\end{table}
\begin{table}
    \small
    \centering
    \begin{tabular}{lrrrrrrrrr}
\toprule
 & \rotatebox{90}{Chronos (Small)} & \rotatebox{90}{ToTo} & \rotatebox{90}{Mantis} & \rotatebox{90}{NuTime} & \rotatebox{90}{Moment (Large)} & \rotatebox{90}{Moment (Base)} & \rotatebox{90}{Moment (Large)} & \rotatebox{90}{DTW (1-NN)} & \rotatebox{90}{DTW (3-NN)} \\
\midrule
UWaveGestureLibraryAll & 0.90 & 0.88 & 0.85 & 0.88 & 0.73 & 0.68 & 0.66 & 0.89 & 0.90 \\
UWaveGestureLibraryX & 0.80 & 0.76 & 0.77 & 0.71 & 0.74 & 0.72 & 0.71 & 0.73 & 0.74 \\
UWaveGestureLibraryY & 0.75 & 0.68 & 0.69 & 0.65 & 0.65 & 0.64 & 0.62 & 0.63 & 0.63 \\
UWaveGestureLibraryZ & 0.72 & 0.71 & 0.73 & 0.65 & 0.66 & 0.67 & 0.66 & 0.66 & 0.67 \\
Wafer & 1.00 & 0.99 & 0.99 & 0.99 & 0.92 & 0.91 & 0.90 & 0.98 & 0.98 \\
Wine & 0.54 & 0.48 & 0.80 & 0.80 & 0.52 & 0.56 & 0.44 & 0.57 & 0.57 \\
WordSynonyms & 0.54 & 0.45 & 0.58 & 0.44 & 0.43 & 0.42 & 0.41 & 0.65 & 0.60 \\
Worms & 0.73 & 0.75 & 0.65 & 0.56 & 0.62 & 0.64 & 0.61 & 0.58 & 0.39 \\
WormsTwoClass & 0.81 & 0.83 & 0.79 & 0.60 & 0.73 & 0.77 & 0.77 & 0.62 & 0.55 \\
Yoga & 0.84 & 0.72 & 0.82 & 0.77 & 0.66 & 0.62 & 0.62 & 0.84 & 0.82 \\
AllGestureWiimoteX & 0.52 & 0.59 & 0.60 & 0.29 & 0.52 & 0.58 & 0.59 & 0.71 & 0.62 \\
AllGestureWiimoteY & 0.50 & 0.60 & 0.59 & 0.30 & 0.54 & 0.64 & 0.61 & 0.68 & 0.61 \\
AllGestureWiimoteZ & 0.51 & 0.56 & 0.63 & 0.28 & 0.48 & 0.56 & 0.53 & 0.70 & 0.64 \\
GestureMidAirD1 & 0.67 & 0.58 & 0.58 & 0.54 & 0.66 & 0.66 & 0.58 & 0.45 & 0.39 \\
GestureMidAirD2 & 0.65 & 0.55 & 0.65 & 0.47 & 0.60 & 0.58 & 0.63 & 0.32 & 0.33 \\
GestureMidAirD3 & 0.41 & 0.32 & 0.33 & 0.33 & 0.37 & 0.39 & 0.34 & 0.18 & 0.15 \\
GesturePebbleZ1 & 0.73 & 0.72 & 0.88 & 0.65 & 0.77 & 0.80 & 0.79 & 0.69 & 0.72 \\
GesturePebbleZ2 & 0.72 & 0.66 & 0.86 & 0.60 & 0.77 & 0.74 & 0.76 & 0.67 & 0.69 \\
PickupGestureWiimoteZ & 0.70 & 0.70 & 0.88 & 0.32 & 0.66 & 0.64 & 0.54 & 0.74 & 0.68 \\
ShakeGestureWiimoteZ & 0.90 & 0.76 & 0.86 & 0.34 & 0.76 & 0.80 & 0.80 & 0.86 & 0.90 \\
DodgerLoopDay & 0.53 & 0.48 & 0.51 & 0.32 & 0.38 & 0.39 & 0.38 & 0.45 & 0.44 \\
DodgerLoopGame & 0.63 & 0.72 & 0.76 & 0.58 & 0.70 & 0.68 & 0.69 & 0.90 & 0.88 \\
DodgerLoopWeekend & 0.94 & 0.83 & 0.95 & 0.89 & 0.93 & 0.87 & 0.90 & 0.95 & 0.96 \\
MelbournePedestrian & 0.92 & 0.87 & 0.91 & 0.97 & 0.59 & 0.68 & 0.64 & 0.88 & 0.88 \\
ArticularyWordRecognition & 0.98 & 0.96 & 0.99 & 0.88 & 0.79 & 0.80 & 0.78 & 0.99 & 0.98 \\
AtrialFibrillation & 0.20 & 0.07 & 0.27 & 0.33 & 0.27 & 0.13 & 0.13 & 0.20 & 0.33 \\
BasicMotions & 1.00 & 1.00 & 1.00 & 0.85 & 0.97 & 0.95 & 0.95 & 0.97 & 0.85 \\
Cricket & 0.90 & 0.99 & 1.00 & 0.72 & 0.46 & 0.44 & 0.47 & 1.00 & 1.00 \\
Epilepsy & 0.99 & 0.99 & 1.00 & 0.88 & 0.98 & 0.98 & 0.98 & 0.96 & 0.95 \\
EthanolConcentration & 0.47 & 0.38 & 0.30 & 0.53 & 0.26 & 0.33 & 0.25 & 0.32 & 0.28 \\
ERing & 0.94 & 0.86 & 0.93 & 0.84 & 0.79 & 0.86 & 0.81 & 0.91 & 0.93 \\
FaceDetection & 0.56 & 0.58 & 0.52 & 0.55 & 0.52 & 0.51 & 0.51 & 0.53 & 0.54 \\
FingerMovements & 0.51 & 0.55 & 0.54 & 0.56 & 0.47 & 0.53 & 0.54 & 0.53 & 0.54 \\
HandMovementDirection & 0.28 & 0.28 & 0.28 & 0.26 & 0.18 & 0.24 & 0.24 & 0.19 & 0.20 \\
Handwriting & 0.17 & 0.15 & 0.33 & 0.13 & 0.16 & 0.15 & 0.12 & 0.61 & 0.50 \\
Heartbeat & 0.76 & 0.75 & 0.79 & 0.72 & 0.72 & 0.70 & 0.71 & 0.72 & 0.73 \\
Libras & 0.89 & 0.84 & 0.88 & 0.78 & 0.44 & 0.56 & 0.49 & 0.87 & 0.86 \\
LSST & 0.55 & 0.56 & 0.61 & 0.43 & 0.52 & 0.51 & 0.49 & 0.55 & 0.56 \\
NATOPS & 0.82 & 0.83 & 0.92 & 0.72 & 0.62 & 0.58 & 0.54 & 0.88 & 0.87 \\
PenDigits & 0.95 & 0.96 & 0.94 & 0.96 & 0.72 & 0.80 & 0.77 & 0.98 & 0.98 \\
PEMS-SF & 0.98 & 0.83 & 0.99 & 0.98 & NaN & NaN & 0.83 & 0.71 & 0.53 \\
PhonemeSpectra & 0.25 & 0.20 & 0.27 & 0.11 & 0.19 & 0.22 & 0.21 & 0.15 & 0.14 \\
RacketSports & 0.88 & 0.82 & 0.92 & 0.83 & 0.58 & 0.59 & 0.51 & 0.80 & 0.83 \\
SelfRegulationSCP1 & 0.76 & 0.82 & 0.80 & 0.75 & 0.69 & 0.62 & 0.62 & 0.77 & 0.83 \\
SelfRegulationSCP2 & 0.52 & 0.52 & 0.46 & 0.45 & 0.50 & 0.48 & 0.49 & 0.54 & 0.49 \\
UWaveGestureLibrary & 0.85 & 0.83 & 0.82 & 0.81 & 0.62 & 0.65 & 0.64 & 0.90 & 0.90 \\
CharacterTrajectories & 0.97 & 0.96 & 0.96 & 0.95 & 0.83 & 0.83 & 0.81 & 0.99 & 0.98 \\
JapaneseVowels & 0.94 & 0.93 & 0.96 & 0.95 & 0.39 & 0.39 & 0.32 & 0.96 & 0.96 \\
SpokenArabicDigits & 0.97 & 0.95 & 0.95 & 0.93 & 0.81 & 0.78 & 0.76 & 0.97 & 0.97 \\
\bottomrule
\end{tabular}

    \caption{Accuracy results of the different models with augmentations on the individual datasets with a Random Forest classifier. (Part 6/6)}
    \label{tab:individual-part6}
\end{table}

\end{document}